\documentclass[journal]{IEEEtran}
\usepackage{cite}
\usepackage{amsmath,amssymb}
\usepackage{graphicx}
\usepackage{booktabs}
\usepackage{multirow}
\usepackage{array}
\usepackage{xcolor}
\usepackage{enumitem}
\usepackage{xspace}
\usepackage{placeins}
\usepackage{adjustbox}
\usepackage[hidelinks]{hyperref}
\usepackage[capitalize]{cleveref}
\crefname{section}{Section}{Sections}
\crefname{table}{Table}{Tables}
\crefname{figure}{Fig.}{Figs.}
\Crefname{figure}{Fig.}{Figs.}
\crefname{equation}{}{}
\crefname{appendix}{Appendix}{Appendices}
\Crefname{appendix}{Appendix}{Appendices}

\newcommand{\pp}{\,pp\xspace}

\renewcommand{\paragraph}[1]{\vspace{0.2em}\noindent\textit{#1}\hspace{0.4em}}

\begin{document}

\title{Size-Matched Nulls Reveal What Block\\ Ablation Measures in MambaVision}

\author{Tahmina~Kousar and Tehseen~Zia%
\thanks{T. Kousar and T. Zia are with the Department of Computer Science, COMSATS University Islamabad, Islamabad, Pakistan (e-mail: SP25-RAI-020@isbstudent.comsats.edu.pk; tehseen.zia@comsats.edu.pk). Corresponding author: Tehseen Zia.}}

\markboth{}{}

\maketitle

\begin{abstract}
Hybrid vision backbones interleave Mamba state-space mixers with self-attention, and a block's role is usually probed by removing it and measuring the damage. We test what that measurement means in MambaVision. On all 50,000 ImageNet validation images and three model sizes, every block ablation is paired with a size-matched null, the same perturbation with its channels shuffled or their signs flipped. In top-1 accuracy and in true-class rank, real and null damage are nearly indistinguishable in absolute terms: pooled over the blocks of MambaVision-T, the content effect lies within $[-0.044,+0.016]$ percentage points (90\% interval). Two attention-only controls show that the test detects content, with real ablations worse than their nulls by up to 4.2 points in DeiT-Small and 1.0 point in Swin-T. Part of MambaVision's near-equivalence reflects how little damage its blocks do (0.11--0.34 points on average, against 1.7 and 2.1). Scaled by damage, its content share is comparable to Swin-T's in accuracy, lower in rank, and far below DeiT-Small's: single-block damage in MambaVision is small and mostly positional. Ablation also hides what blocks carry together: removing all blocks of one type compounds damage to 3.7--8.4 times the sum of single effects, far more than size-matched nulls, including nulls that keep the alignment between the blocks' perturbations; for adjacent pairs, aligned nulls reproduce most of it. Freezing input-dependent selection in all Mamba mixers shows that it accounts for 20--32\% of the Mamba contribution, mainly through the $B$ and $C$ projections. We turn these results into a protocol for causal attribution in hybrid networks.
\end{abstract}


\section{Introduction}
\label{sec:intro}

\IEEEPARstart{H}{ybrid} backbones that combine selective state-space mixers~\cite{gu2023mamba,dao2024transformers} with self-attention~\cite{dosovitskiy2021vit} now match pure Transformers on ImageNet at a lower cost~\cite{hatamizadeh2025mambavision,lieber2024jamba}. Whether the state-space part earns its place is debated: some vision tasks appear not to need it at all~\cite{yu2025mambaout}, and much of its benefit can be explained by its similarity to linear attention~\cite{han2024demystify}. Inside a hybrid model the question becomes \emph{which part does what}: whether the Mamba mixers and the attention mixers take on different roles, and whether the input-dependent selection that defines Mamba is actually used for vision. The standard tool for such questions is causal intervention. One replaces a block's output by a neutral value, for example its mean over a reference set, and reads the change in accuracy as that block's contribution~\cite{vig2020causal,meng2022rome,wang2023ioi}.

This reading assumes that the damage comes from losing what the block computed. It could equally come from the size and location of the perturbation itself: replacing an output by its mean moves the residual stream, and some positions in a network may be fragile to \emph{any} move of that size. Block ablation alone cannot tell these apart, and work on language models shows that localisation by intervention can mislead in exactly this way~\cite{hase2023localization,makelov2024subspace}. The problem is sharper in hybrid vision models, where block type is tied to depth. In MambaVision~\cite{hatamizadeh2025mambavision} the Mamba mixers always come first within a stage and the attention mixers last, so a type difference in ablation damage can just as well be a position difference. The question matters beyond interpretation: ablation scores also drive component pruning and automated circuit discovery~\cite{conmy2023acdc}, so a score that reflects position rather than content can remove the wrong components.

\begin{figure}[t]
  \centering
  \includegraphics[width=\linewidth]{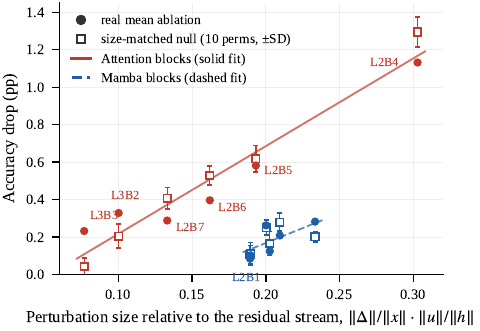}
  \caption{Ablation damage mostly tracks where a perturbation lands rather than what was removed. Accuracy drop of MambaVision-T for the real mean ablation of each mixer block (filled) and for a size-matched shuffled perturbation of the same block (open, mean $\pm$ SD over ten permutations), against the size of the perturbation relative to the residual stream. Real and null damage nearly coincide, and the attention positions (red, solid fit) are several times more sensitive per unit of perturbation than the Mamba positions (blue, dashed fit). 50,000 ImageNet validation images.}
  \label{fig:teaser}
\end{figure}

We separate the two with a \emph{size-matched null}. For every block we add the same perturbation that mean ablation would add, but with its channels permuted, so the perturbation has exactly the same norm and enters at the same place but is no longer aligned with what the block computed. A second null flips the sign of each channel instead, which keeps every channel's magnitude in place. The difference between real and null damage, the content effect $C$, isolates the part of the damage that depends on content. We apply this test, together with group, pairwise and selectivity interventions, to MambaVision-T, -S and -B on all 50,000 ImageNet validation images, and to two attention-only controls: the flat DeiT-Small~\cite{touvron2021deit} and the hierarchical Swin-T~\cite{liu2021swin} (\cref{fig:teaser}).

The paper makes the following contributions.
\begin{enumerate}[leftmargin=*,itemsep=1pt,topsep=2pt]
  \item \textit{Single-block ablation in MambaVision measures little damage, and mostly positional damage.} In top-1 accuracy, real and null damage are rank-correlated ($\rho=0.83$) and the pooled content effect of MambaVision-T lies within $[-0.044,+0.016]$\pp, with similar bounds on -S and -B and against the sign-flip null. On DeiT-Small and Swin-T the same test finds real-worse content effects of up to 4.2\pp and 1.0\pp, so it detects content when it is there. Because MambaVision's blocks do little damage (0.11--0.34\pp on average, against 1.7 and 2.1\pp in the controls), we also scale content effects by damage: MambaVision's content share is comparable to Swin-T's in accuracy, lower in rank, and far below DeiT-Small's. Single-block damage in MambaVision is small and mostly positional; its residual stream is unusually robust to the loss of any one block. In true-class log-probability, small block-specific effects remain ($|C|\lesssim0.1$ nats), but in true-class rank the nulls reproduce the real damage in absolute terms, so these effects are shifts of confidence rather than of evidence.
  \item \textit{Compensation is content-specific at the group level.} Removing all blocks of one type costs far more than the sum of single effects (up to $8.4\times$). Size-matched group nulls compound much less, even when they keep the alignment between the blocks' perturbations, so single-block tests understate what groups of blocks carry. For adjacent pairs, aligned nulls reproduce most of the compounding, so the natural unit of attribution is a group of same-type blocks.
  \item \textit{Selectivity is real but collective, and it lives in $B$ and $C$.} Per block, freezing the input-dependent scan parameters costs at most 0.23\pp. Frozen in all Mamba mixers at once, image-dependence alone accounts for 20--32\% of the Mamba contribution across three model sizes, while the step size $\Delta$ acts mostly as a fixed position profile. A positive control on a language Mamba validates the freezing. Without selection the model becomes more confident while making more errors, and the frozen scan is an exact long convolution whose long-lag taps still matter.
  \item \textit{A protocol for causal attribution in hybrid networks.} We define the nulls, the statistics and the reference-set construction precisely (\cref{sec:method}) and condense the results into a step-by-step checklist with a worked example (\cref{sec:checklist}). Most steps change at least one conclusion in this paper, as the worked example shows.
\end{enumerate}

\Cref{sec:related} reviews related work and \cref{sec:method} describes the method. \Cref{sec:null,sec:compensation,sec:selectivity} report the three main findings, and \cref{sec:robust} tests them under a different input resolution and under corruptions. \Cref{sec:discussion} discusses what the results mean and where they stop, \cref{sec:checklist} gives the protocol, and the appendices give per-block results for all five models, channel statistics, the SSM-versus-gate pathway analysis, a sampling audit, the full freezing grid, the interaction range of each block and per-block results under corruption.

\section{Related Work}
\label{sec:related}

\subsection{Mamba in Vision and Its Interpretation}
Mamba~\cite{gu2023mamba} makes the step size $\Delta$ and the projections $B$ and $C$ of a structured state-space model~\cite{gu2022s4} functions of the input, which allows it to select and copy information~\cite{jelassi2024copying}. Vision backbones built from Mamba blocks scan image patches along one or more orders~\cite{zhu2024vim,liu2024vmamba}, and hybrids such as MambaVision~\cite{hatamizadeh2025mambavision} place Mamba mixers before self-attention within each stage. Whether the recurrent part is needed is contested. MambaOut~\cite{yu2025mambaout} removes the state-space model from the Mamba block and matches the original on ImageNet classification, arguing that the long-sequence and autoregressive properties Mamba is built for are not what image classification requires. Han et al.~\cite{han2024demystify} rewrite the selective scan as a variant of linear attention and attribute most of Mamba's success in vision to its forget gate and block design. Both studies compare architectures trained from scratch; we ask a complementary question about a trained hybrid, namely how much of what its Mamba blocks contribute depends on input-dependent selection.

Interpretation methods have followed. The scan can be written as an implicit attention matrix~\cite{ali2025hidden,dao2024transformers}, which gives Mamba models attention-style attribution maps. MambaLRP~\cite{jafari2024mambalrp} adapts layer-wise relevance propagation to selective state-space models so that relevance is conserved through the scan, and patching methods developed for Transformers transfer to Mamba language models~\cite{sharma2024mambaedit}. Vision Mamba models develop high-norm artefact tokens that registers remove~\cite{wang2025mambareg}, mirroring the register tokens of ViTs~\cite{darcet2024registers}. These methods attribute a prediction to inputs or tokens. We attribute it to blocks and to the selection mechanism, by intervention, and test whether such interventions measure what they appear to measure.

\subsection{Ablation, Patching and Their Pitfalls}
Intervening on internal activations and measuring the change in output is the backbone of mechanistic interpretability, from causal mediation analysis~\cite{vig2020causal} and causal abstraction~\cite{geiger2021causal} to activation patching for localising facts and circuits~\cite{meng2022rome,wang2023ioi}. ACDC~\cite{conmy2023acdc} automates the search for circuits by pruning edges whose patching changes a task metric by less than a threshold. Causal scrubbing~\cite{chan2022scrubbing} tests a hypothesis by resampling activations from other inputs that the hypothesis says should be equivalent, which made resample ablation a common alternative to mean ablation. Recent work shows that the outcome of such interventions depends on choices that are often left implicit: the replacement value (zero, mean, resampled or optimised~\cite{li2024optimal}), the metric (probability, logit difference or accuracy~\cite{zhang2024patching}) and the reference distribution~\cite{heimersheim2024patching}.

Several studies show that a successful intervention need not mean what it seems to. Hase et al.~\cite{hase2023localization} find that where causal tracing localises a fact in a language model does not predict which layer is best to edit. Makelov et al.~\cite{makelov2024subspace} show that patching a subspace can change behaviour by activating a dormant pathway rather than the one that carries the feature. Head- and layer-removal studies in Transformers report that many components can be removed with little loss~\cite{michel2019heads,voita2019heads}, and removal-based attribution in vision needs care because removal itself shifts the input distribution~\cite{hooker2019roar}. Our size-matched null addresses the same problem from a different side: it asks how much of an intervention's effect would follow from any perturbation of the same size at the same place.

\subsection{Random Baselines and Controls}
A finding about a representation is informative only against a baseline that lacks the property in question. TCAV~\cite{kim2018tcav} tests each concept direction against directions learned from random image sets and keeps only concepts that beat them. Control tasks~\cite{hewitt2019control} compare a probe's accuracy on a linguistic task with its accuracy on a task with random labels, so that a probe's own capacity is not mistaken for structure in the representation. Our nulls carry this logic to block ablation: they keep the norm, position and, for the sign-flip null, the per-channel magnitude of the ablation, and remove only its relation to what the block computed.

\subsection{Self-Repair and Redundancy}
Residual networks behave like ensembles of shallower paths, so removing one block hurts little~\cite{veit2016residual}. In language models, later components partly restore the output when an earlier one is ablated, an effect called self-repair or the Hydra effect~\cite{mcgrath2023hydra,rushing2024selfrepair}. We measure the analogous effect in a hybrid vision model with group and pairwise ablation, and show with size-matched nulls that the compensation depends on block content rather than on perturbation size.

\section{Method}
\label{sec:method}

\subsection{Models and Data}
MambaVision~\cite{hatamizadeh2025mambavision} has four stages; the first two are convolutional and the last two are hybrid. We name the mixer blocks of the hybrid stages L2B$j$ and L3B$j$ (stage index from zero). In MambaVision-T, L2B0--L2B3, L3B0 and L3B1 are Mamba mixers and L2B4--L2B7, L3B2 and L3B3 are attention mixers; MambaVision-S has 7 Mamba and 5 attention mixers and MambaVision-B has 8 and 7. Within each stage all Mamba mixers precede all attention mixers. Two attention-only models serve as controls: DeiT-Small~\cite{touvron2021deit}, a flat ViT with 12 attention layers, and Swin-T~\cite{liu2021swin}, a hierarchical Transformer with 12 windowed-attention blocks in four stages (depths 2-2-6-2) whose last two stages run at the same $14\times14$ and $7\times7$ token resolutions as the hybrid stages of MambaVision. For both we ablate the output of each attention module, including its output projection, and name Swin blocks S$i$B$j$. We use the released ImageNet-1k checkpoints of all four models.

All experiments use the 50,000 ImageNet-1k validation images~\cite{russakovsky2015imagenet} in one fixed random order (seed 42). MambaVision models use the evaluation transform of their released configuration (for MambaVision-T, the short side is resized to 224 pixels and centre-cropped to $224\times224$), with ImageNet mean and standard deviation. The preprocessed images of each model are cached once and reused by every experiment on that model. Baseline top-1 accuracy is 82.19\% (T), 83.25\% (S), 84.13\% (B), 79.85\% (DeiT-Small) and 81.37\% (Swin-T), matching the published values. DeiT-Small and Swin-T share one image cache (bicubic resize of the short side to 248 pixels, centre crop $224\times224$). Inference is in float32.

\begin{figure}[t]
  \centering
  \includegraphics[width=\linewidth]{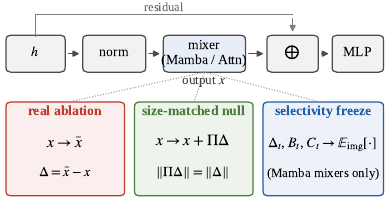}
  \caption{Interventions. All interventions act on the output $x$ of one block's mixer; the residual path, normalisation and MLP are untouched. The size-matched null adds the real ablation's perturbation $\Delta$ with its channels permuted ($\Pi$), so $\|\Pi\Delta\|=\|\Delta\|$ at every token.}
  \label{fig:method}
\end{figure}

\subsection{Reference Sets}
Mean ablation and selectivity freezing replace activations by statistics of a reference set. For every model the reference set is the first 256 images of the fixed order. On these images we record, for each mixer, the per-channel mean of its output over images and tokens, and for each Mamba mixer the reference values of $\Delta_t$, $B_t$ and $C_t$ (below). The pathway analysis of \cref{app:pathway} uses the first 2,000 images. Reference images are part of the evaluation set; re-scoring every effect on the 48,000 images that no reference set touches changes no effect by more than 0.024\pp (\cref{app:audit}). For the 384-pixel experiments the statistics are recomputed at that resolution (\cref{sec:robust}).

\subsection{Interventions}
\label{sec:interventions}
\paragraph{Mean ablation.} The mixer output $x\in\mathbb{R}^{L\times D}$ of block $b$ ($L$ tokens, $D$ channels) is replaced by its reference mean $\bar{x}$, which adds the perturbation $\Delta=\bar{x}-x$ to the residual stream.

\paragraph{Shuffled null.} The null adds $\Pi\Delta$ instead, where $\Pi$ permutes the $D$ channels with a permutation that is fixed for the block and shared by all images and tokens. This keeps the norm of the perturbation at every token and its point of entry, and removes its relation to what the block computed. Permutation 0 is drawn with seed 42 in block order; permutations $j=1,\dots,9$ of block index $b$ use seed $1000+100b+j$. For every model we average the null over ten permutations per image.

\paragraph{Sign-flip null.} Shuffling can move large values into channels that are normally small. The sign-flip null adds $S\Delta$, with $S=\mathrm{diag}(s_1,\dots,s_D)$ and $s_d\in\{\pm1\}$ drawn once per block (seed $2000+100b+j$, three seeds), which keeps the magnitude of every channel in place. In the channels with $s_d=+1$, about half of them, this null equals the real ablation, so it departs from the real perturbation in only half of the channels and is a weaker null than shuffling; we use it as a check on the shuffled null, not as a replacement. It is compared with shuffled permutations 0--2 on the same images in a single forward pass.

\paragraph{Resample ablation.} As an in-distribution alternative to both, the mixer output of each image is replaced by the output of the same block on an image drawn at random (seed $777+b$) from a bank of 256 evaluation images (selected with seed 777); three seeds per block ($777+b$, $1777+b$, $2777+b$), compared with the real ablation on the same images, so the replacement is a real activation of the block, but for the wrong image~\cite{chan2022scrubbing}.

\paragraph{Group and pair nulls.} Group interventions ablate all Mamba or all attention mixers of a model at once; pair interventions ablate two blocks. The corresponding nulls apply the shuffled null to the same blocks at once, each block with its own permutation.

\paragraph{Selectivity freezing.} The input-dependent scan parameters of a Mamba mixer, $\Delta_t$ (before the softplus), $B_t$ and $C_t$, are replaced by reference values. In \emph{mean mode} the reference is averaged over images and tokens, which removes both image- and position-dependence; in \emph{position mode} it is averaged over images only, which removes image-dependence and keeps a fixed profile over token positions. We freeze each tensor alone and all three together, in single mixers and in all Mamba mixers jointly.

\paragraph{Explicit kernel.} With $\Delta$, $B$ and $C$ frozen in mean mode, the scan is a linear time-invariant filter. Each channel $d$ applies the causal kernel
\begin{equation}
  K_d[k]=\sum_{n} \bar{C}_n\exp\!\big(k\,\bar\Delta_d A_{dn}\big)\,\bar\Delta_d \bar{B}_n,\qquad k=0,\dots,L-1,
\end{equation}
along the token sequence, which we implement as a depthwise convolution. Truncation keeps the first $k$ taps, either as they are, rescaled to the original DC gain, or with the removed tail added to the last kept tap.

\subsection{Metrics}
For each image we record whether the top-1 prediction is correct, the true-class log-probability drop $\log p_{\text{base}}(y)-\log p_{\text{int}}(y)$, $\mathrm{KL}(p_{\text{base}}\,\|\,p_{\text{int}})$, and the rank of the true class among the 1,000 logits (0 for a correct prediction). The rank does not change when the logits are rescaled, so it separates evidence from confidence; we use $\log(1+\text{rank})$. For calibration we report the mean maximum softmax probability, the predictive entropy, the negative log-likelihood and the expected calibration error (ECE) with 15 equal-width bins~\cite{guo2017calibration}. A no-op pass through the full pipeline gives $\mathrm{KL}=8.0\times10^{-7}$ and identical predictions.

\subsection{Content Effect and Statistical Procedure}
\label{sec:stats}
For block $b$ the content effect in accuracy is
\begin{equation}
  C_b = \mathbb{E}_i\big[\mathbf{1}_{\text{null}}(i) - \mathbf{1}_{\text{real}}(i)\big]\times 100,
\end{equation}
where $\mathbf{1}(i)$ indicates a correct top-1 prediction for image $i$ under each intervention and the null indicator is averaged over permutations. $C_b>0$ means the real ablation hurts more than a perturbation of the same size, so the block's content matters beyond magnitude. The same definition applies to the other metrics, with the sign chosen so that $C>0$ always means that the real ablation is worse.

All intervals are paired percentile bootstraps over images with 2,000 resamples; because every intervention is evaluated on the same images as the unmodified model, the bootstrap resamples images jointly across conditions. Single ablations are compared with the baseline by McNemar's test. Tests across the blocks of a model are corrected with the Holm procedure~\cite{holm1979}. To summarise a model we average $C$ over its blocks for each image and bootstrap this per-image mean.

A non-significant $C$ does not show that real and null damage are equal, so we test equivalence~\cite{schuirmann1987tost}: the pooled effect is declared equivalent to zero within a margin $X$ when its 90\% interval lies inside $[-X,+X]$. We fixed $X=0.10$\pp after the single-permutation pilot and before the multi-permutation runs. Two independent quantities support this choice. First, the null has its own spread: across the ten permutations, the null damage of a single block of MambaVision-T has a standard deviation of 0.03--0.08\pp (median 0.05\pp) and a range of 0.10--0.24\pp (median 0.17\pp). A content effect smaller than $X$ is therefore smaller than the typical difference between two equally valid draws of the null on one block; pooled over blocks, a single permutation already moves $C$ between $-0.049$ and $+0.014$\pp. Second, $X$ is 30\% of the mean single-block damage of MambaVision-T (0.34\pp), and it exceeds the single-block damage itself only for L2B1 (0.08\pp). A third anchor, the variation of these effects between training seeds, is not available because one checkpoint per model size has been released (\cref{sec:discussion}).

An absolute margin favours models whose blocks do little damage, because a block that costs 0.1\pp cannot have a content effect much larger than that. We therefore also scale content effects by damage. With $d_b$ the accuracy drop under the real ablation of block $b$ and $v_b$ the sampling variance of $C_b$, the \emph{content share} $R$ of a model is
\begin{equation}
  R=\sqrt{\frac{\sum_b \max\!\big(C_b^2-v_b,\,0\big)}{\sum_b d_b^2}},
\end{equation}
the root-mean-square content effect relative to the root-mean-square damage, corrected for the noise that inflates $C_b^2$. $R=0$ means that nulls reproduce the damage of every block, and $R=1$ that the content effects are as large as the damage itself. Unlike the median of $|C_b|/|d_b|$, $R$ is not dominated by blocks whose damage is close to zero. Its interval is a bootstrap over images (2,000 resamples, $C_b^{*2}-2v_b$ in each replicate). Confidence intervals resample images; the null is averaged over ten permutations, and treating the permutations as a second source of variance widens the pooled interval of MambaVision-T by 4\% and per-block intervals by at most 11\%, which changes no conclusion. As a reference for $R$ we also report the noise floor $\sqrt{\sum_b v_b/\sum_b d_b^2}$, the uncorrected value that sampling noise alone would produce. The same quantity is computed for $\log(1+\text{rank})$. As a relative equivalence test we ask whether the one-sided 95\% upper bound of $R$ lies below $0.5$, that is, whether content effects account for at most a quarter of the squared damage. This margin was set after the absolute analysis, in response to review, and should be read as descriptive.

\subsection{Compute}
All experiments ran on NVIDIA Tesla T4 GPUs (one or two per run). Evaluating MambaVision-T on the 50,000 cached images takes about 2 minutes, so one block with the real ablation, three shuffled and three sign-flip nulls, run as seven copies of each batch after a shared prefix, takes 14 minutes. Interventions are implemented as forward hooks on the released PyTorch models, and every result is checked against an earlier run where one exists (for example, the shuffled nulls of \cref{sec:null} reproduce the earlier ten-permutation run on 100\% of images). The code is available (see Code and Data Availability).

\section{Block ablation versus a size-matched null}
\label{sec:null}

\begin{table*}[t]
\centering\small\setlength{\tabcolsep}{4pt}
\caption{Real vs Size-Matched Null, Single Blocks. $C$: Content Effect (Null Minus Real Accuracy Change). Pooled $C$: Per-Image Mean Over Blocks With 95\% Bootstrap CI. $\rho$: Spearman Correlation of $|\text{real}|$ and $|\text{null}|$ Damage Over Blocks. Holm: Blocks With a Significant $C$ After Correction. Mean Damage: Mean $|d_b|$. Content Share $R$: Noise-Corrected RMS Content Effect Relative to RMS Damage (\cref{sec:stats}), in Accuracy and in $\log(1+\text{rank})$, With 95\% CI. DeiT-Small (Flat) and Swin-T (Hierarchical) Are Attention-Only Controls; Swin-T Stages 2--3 Run at the Resolutions of MambaVision's Hybrid Stages (Holm Correction Over All 12 Swin-T Blocks)}
\label{tab:null}
\begin{adjustbox}{max width=\textwidth}\begin{tabular}{@{}lccccccc@{}}
\toprule
Model & $\rho$ & pooled $C$ (pp) & max $|C|$ & Holm: real / null worse & mean damage (pp) & $R$, accuracy & $R$, rank\\
\midrule
MV-T & 0.83 & $-$0.016 [$-$0.049, $+$0.024] & 0.19 & L3B3 / none & 0.34 & 0.20 [0.06, 0.31] & 0.06 [0.00, 0.12]\\
MV-S & 0.64 & $-$0.010 [$-$0.045, $+$0.024] & 0.33 & L3B3 / L2B4 & 0.21 & 0.42 [0.25, 0.61] & 0.16 [0.08, 0.22]\\
MV-B & 0.55 & $+$0.037 [$+$0.011, $+$0.062] & 0.13 & L3B3 / none & 0.11 & 0.25 [0.00, 0.46] & 0.39 [0.28, 0.48]\\
\midrule
Swin-T & 0.68 & $+$0.071 [$+$0.012, $+$0.129] & 1.25 & 6 blocks / 2 blocks & 1.66 & 0.24 [0.22, 0.26] & 0.37 [0.36, 0.39]\\
\quad stages 2--3 & -- & -- & 1.25 & 3 blocks / 1 block & 0.98 & 0.44 [0.38, 0.51] & 0.94 [0.88, 1.01]\\
DeiT-S & 0.50 & $-$0.108 [$-$0.163, $-$0.050] & 8.75 & 5 layers / 2 layers & 2.10 & 0.91 [0.88, 0.94] & 0.97 [0.95, 0.99]\\
\bottomrule
\end{tabular}\end{adjustbox}
\end{table*}

\noindent\textit{In short:} single blocks of MambaVision do little damage, and a perturbation of the same size at the same place does nearly the same damage; relative to that small damage, the content-specific part is about as large as in a hierarchical Transformer.

\paragraph{Accuracy: damage tracks the null.}
Every single-block mean ablation of MambaVision-T lowers accuracy, from 0.08\pp (L2B1) to 1.13\pp (L2B4), and the attention mixers cost more on average (0.49 vs.\ 0.18\pp). A size-matched shuffled perturbation of the same block reproduces this pattern almost exactly (\cref{fig:teaser}). Across blocks the two are rank-correlated ($\rho=0.83$, \cref{tab:null}), and after Holm correction only one block, L3B3, is more damaged by its real ablation than by its null ($C=+0.19$\pp). Pooled over the 12 blocks, $C=-0.016$\pp $[-0.049,+0.024]$, and the 90\% interval $[-0.044,+0.016]$\pp lies well inside the margin of $\pm0.10$\pp. The ten permutations give the null its own spread (SD 0.03--0.08\pp per block), so the bound does not rest on one lucky draw. Nor does it rest on the choice of null: against three sign-flip nulls on the same images, pooled $C=-0.005$\pp $[-0.042,+0.037]$ (90\%: $[-0.036,+0.030]$), and real and sign-flip damage are rank-correlated as well ($\rho=0.75$). MambaVision-S, also with ten permutations, gives the same picture ($\rho=0.64$, pooled $C=-0.010$\pp, 90\% interval $[-0.038,+0.019]$), and its significant blocks are the same as with one permutation. In MambaVision-B single-block effects are small (mean $|\Delta|$ 0.06\pp for Mamba and 0.15\pp for attention), so the rank correlation is weaker ($\rho=0.55$); with ten permutations its pooled $C$ is $+0.037$\pp $[+0.011,+0.062]$ (90\%: $[+0.016,+0.058]$), inside the margin but not centred on zero. B also shows how a single permutation can mislead: L2B6, L2B7 and L3B0 looked content-specific with one permutation ($C=+0.12$ to $+0.14$\pp), but with ten their effects shrink to $+0.00$ to $+0.09$\pp and only L3B3 remains significant ($+0.13$\pp).

\paragraph{The null can detect content.}
A null that always agrees with the real ablation would be useless. Run on DeiT-Small with the same code and ten null permutations per layer, the test separates real and null damage by up to 8.7\pp in a single layer and finds significant content effects in seven of twelve layers, five real-worse and two null-worse, and in the rank of the true class in eight (\cref{tab:null}). DeiT-Small differs from MambaVision in more than its mixers, however: it is flat, whereas MambaVision is hierarchical, with stages of falling resolution and windowed attention. Swin-T~\cite{liu2021swin} shares this hierarchical design but has no Mamba mixers. With ten null permutations per block, its 12 attention blocks show the same picture as DeiT-Small. Eight blocks have a significant content effect after correction, six real-worse and two null-worse, with $|C|$ up to 1.25\pp, and real and null damage are less tightly coupled than in MambaVision-T ($\rho=0.68$). The effects are not confined to the high-resolution early stages: in the last two stages of Swin-T, which run at the same $14\times14$ and $7\times7$ resolutions as the hybrid stages of MambaVision, four of eight blocks have significant accuracy effects (up to 1.25\pp) and seven of eight significant rank effects. Unlike in MambaVision, the content effects of Swin-T appear in the rank of the true class in eleven of twelve blocks ($|C|$ up to 0.051, against at most 0.002 in MambaVision-T), so they change the ordering of classes and not only confidence. Because these block effects have both signs, they partly cancel when pooled: the pooled accuracy effect of Swin-T, $+0.07$\pp $[+0.01,+0.13]$, is small and its 90\% interval $[+0.02,+0.12]$ only just leaves the equivalence margin, and the pooled rank effect is not significant. A pooled bound alone therefore does not separate MambaVision from the controls. Block by block, MambaVision looks different in absolute terms: at most two of 12--15 blocks per model have a significant content effect, with $|C|\le0.33$\pp (0.19\pp in MambaVision-T), against seven or eight blocks and up to 1.25 and 8.75\pp in the two controls, and in all three MambaVision models the rank effects stay at $|C|\le0.003$. Ten permutations also mattered for Swin-T: the single-permutation estimate for its first block, S0B0, was $+1.65$\pp, but that block's null varies strongly between permutations (SD 1.32\pp, the largest of any block in any model) and the ten-permutation estimate is $+0.32$\pp; no other Swin block changed by more than 0.27\pp. For DeiT-Small, going from one to ten permutations changed per-layer $C$ by at most 0.64\pp and added one significant layer.

\paragraph{Relative to damage, MambaVision resembles hierarchical attention.}
Absolute content effects are small in MambaVision partly because its damage is small: the mean single-block damage is 0.34, 0.21 and 0.11\pp in MambaVision-T, -S and -B, against 1.66\pp in Swin-T and 2.10\pp in DeiT-Small. A block that costs 0.1\pp cannot carry a content effect of 1\pp, so the absolute margin of $\pm0.10$\pp is easier to meet in MambaVision than in the controls. Scaled by damage (\cref{tab:null}), the picture changes. In accuracy the content share $R$ of MambaVision is 0.20 [0.06, 0.31] for T, 0.42 [0.25, 0.61] for S and 0.25 [0.00, 0.46] for B, against 0.24 [0.22, 0.26] for Swin-T and 0.91 [0.88, 0.94] for DeiT-Small. MambaVision is thus far below the flat control but comparable to the hierarchical one; against the stages of Swin-T that run at MambaVision's hybrid resolutions ($R=0.44$ [0.38, 0.51]), only MambaVision-T is clearly lower. In the rank of the true class all three MambaVision models lie below Swin-T's matched stages (0.06, 0.16 and 0.39 against 0.94) and below DeiT-Small (0.97), and T and S also lie below Swin-T as a whole (0.37). Because MambaVision's damage is small, its estimates are noisier: the noise floor is 0.13, 0.18 and 0.32 for T, S and B (0.03--0.07 in the controls), so the accuracy share of MambaVision-B cannot be separated from noise. The median of $|C_b|/|d_b|$, a cruder summary, does not separate the five models (0.23--0.44), because blocks with near-zero damage give unstable ratios. Against the relative margin $R<0.5$ (one-sided 95\% bound), MambaVision-T and -B and Swin-T as a whole are equivalent in accuracy, MambaVision-S and DeiT-Small are not, and Swin-T's matched stages sit at the boundary (0.497); in rank, all three MambaVision models and Swin-T as a whole are equivalent, while Swin-T's matched stages and DeiT-Small are not. The accurate reading is therefore not that MambaVision lacks content effects that hierarchical models have. Single-block damage in MambaVision is small and, in absolute terms, mostly positional: its residual stream is unusually robust to the removal of any one block. Relative to that small damage, content effects in accuracy are about as large as in Swin-T and much smaller than in DeiT-Small; in rank they are smaller than in either control.

\paragraph{One block per model, at different positions.}
The only block that is significantly real-worse in every MambaVision model is called L3B3, but it sits in different places: in T it is the second and last attention mixer of stage 3, while in S and B it is the first of two. In T, the first attention mixer of stage 3 (L3B2) is real-worse by its confidence interval against the shuffled null ($C=+0.12$\pp $[+0.01,+0.23]$) and significantly so against the sign-flip null ($+0.20$\pp $[+0.07,+0.31]$, Holm-corrected), as is L3B3 ($+0.22$\pp). The robust statement is that a stage-3 attention mixer carries a small content-specific effect in each model.

\paragraph{Positional fragility.}
If damage is not about content, what sets it? \Cref{fig:teaser} plots damage against the size of the perturbation relative to the residual stream. The attention positions fall on one line (null: $5.4$\pp per unit size, $r=1.00$; real: $4.0$\pp, $r=0.97$), while the six Mamba positions cluster at sizes of 0.19--0.23 with damage of 0.08--0.28\pp. At matched size ($\approx0.19$), perturbing L2B5 costs 0.58\pp and perturbing L2B1 costs 0.08\pp. The residual stream is several times more sensitive at the attention positions than at the Mamba positions, whatever the perturbation carries. Because type and depth coincide in this architecture, the experiment cannot say whether the sensitivity belongs to the attention type or to the positions it occupies.

\begin{table*}[t]
\centering\footnotesize\setlength{\tabcolsep}{4pt}
\caption{Content effect of single blocks in three metrics and against two nulls. MambaVision-T, 50,000 images. $C>0$: the real ablation is worse than the null. Accuracy against the shuffled null uses ten permutations, all other columns three shuffled permutations or three sign-flip seeds on the same images. Log-probability: null minus real $\log p(y)$ (nats); all its 95\% CIs are narrower than $\pm0.003$. Rank: real minus null $\log(1+\text{rank of the true class})$, $\times10^{3}$. 95\% paired bootstrap CIs; $^{*}$ Holm $p<0.05$ over 12 blocks.}
\label{tab:metric}
\begin{tabular}{@{}llcccccc@{}}
\toprule
& & \multicolumn{2}{c}{accuracy $C$ (pp)} & \multicolumn{2}{c}{log-prob.\ $C$ (nats)} & \multicolumn{2}{c}{rank $C$ ($\times10^{3}$)}\\
\cmidrule(lr){3-4}\cmidrule(lr){5-6}\cmidrule(l){7-8}
Block & type & shuffled & sign-flip & shuffled & sign-flip & shuffled & sign-flip\\
\midrule
L2B0 & M & $+$0.081 [$-$0.024, $+$0.186] & $+$0.037 [$-$0.075, $+$0.154] & $+$0.041$^{*}$ & $+$0.045$^{*}$ & $+$1 & $+$1\\
L2B1 & M & $-$0.023 [$-$0.111, $+$0.068] & $-$0.053 [$-$0.149, $+$0.041] & $+$0.017$^{*}$ & $+$0.021$^{*}$ & 0 & $+$1\\
L2B2 & M & $-$0.007 [$-$0.090, $+$0.071] & $+$0.039 [$-$0.051, $+$0.128] & $+$0.019$^{*}$ & $+$0.022$^{*}$ & 0 & $+$1\\
L2B3 & M & $-$0.041 [$-$0.122, $+$0.041] & $-$0.027 [$-$0.119, $+$0.067] & $+$0.012$^{*}$ & $+$0.009$^{*}$ & 0 & 0\\
L2B4 & A & $-$0.163 [$-$0.305, $-$0.003] & $-$0.213 [$-$0.375, $-$0.039] & $-$0.030$^{*}$ & $-$0.037$^{*}$ & 0 & $+$1\\
L2B5 & A & $-$0.036 [$-$0.161, $+$0.089] & $-$0.007 [$-$0.138, $+$0.130] & $+$0.005$^{*}$ & $+$0.004$^{*}$ & 0 & $+$1\\
L2B6 & A & $-$0.132 [$-$0.246, $-$0.012] & $-$0.159 [$-$0.292, $-$0.023] & $+$0.001 & $-$0.010$^{*}$ & 0 & 0\\
L2B7 & A & $-$0.119 [$-$0.237, $-$0.005] & $-$0.106 [$-$0.238, $+$0.021] & $-$0.019$^{*}$ & $-$0.015$^{*}$ & 0 & 0\\
L3B0 & M & $+$0.012 [$-$0.094, $+$0.117] & $+$0.025 [$-$0.095, $+$0.138] & $+$0.004$^{*}$ & $-$0.018$^{*}$ & $-$1 & 0\\
L3B1 & M & $-$0.072 [$-$0.185, $+$0.041] & $-$0.011 [$-$0.128, $+$0.117] & $-$0.010$^{*}$ & $+$0.011$^{*}$ & $-$1 & 0\\
L3B2 & A & $+$0.124 [$+$0.007, $+$0.234] & $+$0.198$^{*}$ [$+$0.074, $+$0.313] & $-$0.087$^{*}$ & $-$0.103$^{*}$ & $-$2$^{*}$ & 0\\
L3B3 & A & $+$0.190$^{*}$ [$+$0.097, $+$0.285] & $+$0.221$^{*}$ [$+$0.124, $+$0.324] & $-$0.066$^{*}$ & $-$0.044$^{*}$ & 0 & $+$1\\
\midrule
pooled & & $-$0.016 [$-$0.049, $+$0.024] & $-$0.005 [$-$0.042, $+$0.037] & $-$0.0096 & $-$0.0095 & $-$0.1 & $+$0.5\\
\bottomrule
\end{tabular}
\end{table*}

\paragraph{The conclusion is metric-specific, and rank separates confidence from evidence.}
Accuracy only registers a change when the top-1 class flips. In the true-class log-probability, real and null ablations do differ, and with per-image intervals almost every block is significant (\cref{tab:metric}). The early Mamba mixers L2B0--L2B3 lower $\log p(y)$ more under real ablation ($+0.009$ to $+0.045$ nats against either null), whereas the stage-3 attention mixers L3B2 and L3B3 lower it more under the null ($-0.04$ to $-0.10$ nats). The pooled log-probability effect is the same against both nulls ($-0.0096$ and $-0.0095$ nats). Accuracy and log-probability are unrelated at the block level (Spearman $\rho=0.06$), and L3B3 is real-worse in accuracy but null-worse in log-probability. The rank of the true class resolves this. It does not change when the logits are rescaled. Real ablations raise $\log(1+\text{rank})$ by 0.002--0.024 (mean 0.006, baseline 0.23), and both nulls reproduce this block by block (L2B4: 0.0238 real, 0.0235 shuffled, 0.0229 sign-flip). Pooled, the content effect in rank is $-0.0001$ $[-0.0005,+0.0003]$ against the shuffled and $+0.0005$ $[+0.0000,+0.0009]$ against the sign-flip null, 2\% and 8\% of the mean real damage. The log-probability differences are therefore shifts of confidence that leave the ordering of classes intact; several interventions, including ablating L3B3 and freezing the scan (\cref{sec:selectivity}), raise $p(y)$ on average while lowering accuracy. The precise form of our first finding is: \emph{in top-1 accuracy and in true-class rank, single-block damage is bounded close to that of a size-matched null; in log-probability, blocks shift confidence in block-specific directions ($|C|\lesssim0.1$ nats).}

\paragraph{Is the shuffled null itself off-distribution?}
Shuffling keeps the norm of the perturbation but moves large values into channels that are normally small, which can be harsh in networks with outlier channels~\cite{sun2024massive,dettmers2022llmint8}. In DeiT layer~0, for instance, the null costs 10.1\pp (mean of ten permutations) against 1.4\pp for the real ablation. If this inflated the null in MambaVision, the near-equivalence could be an artefact. The sign-flip null rules this out. It keeps every channel's magnitude in place, yet its damage tracks the shuffled null closely across blocks ($\rho=0.91$), it yields the same pooled bound and the same pooled log-probability effect, and the block that is most null-worse under shuffling, L2B4 ($C=-0.16$\pp with ten permutations and $-0.25$\pp with the three run alongside the sign-flip null), stays null-worse under sign flips ($-0.21$\pp; none survives correction). Resample ablation, which replaces a block's output by its output on another image and so stays on the block's own activation distribution, points the same way. With three resample seeds per block on all 50,000 images, its damage ranks the 12 blocks almost exactly as the real ablation does (Spearman $\rho=0.96$), with L2B4 again the most damaged block, and it is never significantly milder than the real ablation: seven blocks are more damaged by resampling after correction and none less (pooled $C=-0.18$\pp $[-0.21,-0.16]$). This is what the fragility account predicts. A resampled activation carries another image's content, and because it differs from the current activation by the difference of two samples rather than by one sample's deviation from the mean, its perturbation is about $\sqrt{2}$ times larger than that of mean ablation; it should therefore hurt more at the same positions. The in-distribution alternative thus confirms rather than removes the positional pattern. The only notable change is that the sign-flip null is milder at the stage-3 attention mixers, which makes their small content effect clearer. Where the two nulls differ is predictable from channel statistics on the reference images (\cref{app:channels}). In the stage-2 attention mixers the output has outlier channels (max/median channel RMS 5--9), but the ablation perturbation does not (1.7--2.2), so shuffling moves little mass. In L2B3 and L3B0--L3B3 the perturbation itself is concentrated (max/median 7--13; the top 1\% of channels carry 11--29\% of its energy), and in four of these five blocks the shuffled null is harsher than the sign-flip null.

\section{Compensation is content-specific at the group level}
\label{sec:compensation}

\begin{figure*}[t]
  \centering
  \includegraphics[width=\linewidth]{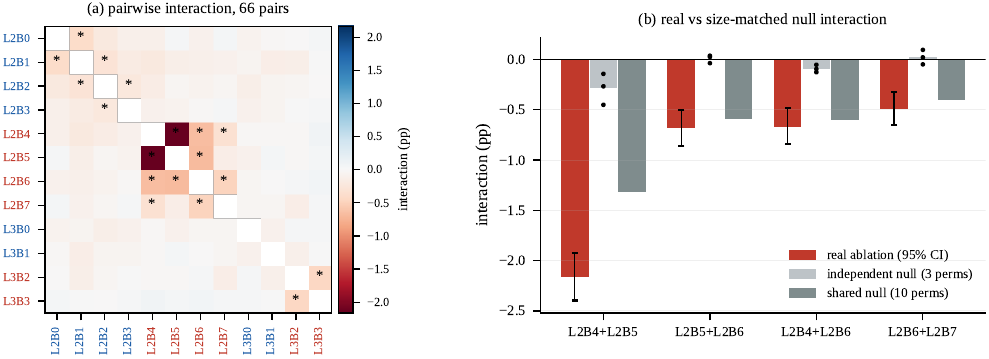}
  \caption{Pairwise ablation. (a) Interaction $\Delta(a{+}b)-\Delta(a)-\Delta(b)$ for all 66 pairs of MambaVision-T mixers (blue labels: Mamba, red: attention; $*$: Holm $p<0.05$ over 66 tests). Negative values mean that removing both costs more than the sum of removing each. (b) For the four strongest pairs, the real interaction (95\% CI) and the interaction under size-matched shuffled perturbations of both blocks, with an independent permutation per block (three permutations, dots) or one shared permutation for both blocks (ten permutations). Only L2B4+L2B5 compounds more than its shared null.}
  \label{fig:pairs}
\end{figure*}

\begin{table}[t]
\centering\footnotesize\setlength{\tabcolsep}{2.6pt}
\caption{Group Ablation Compounds Far More Than Group Nulls. Accuracy Change (pp) When All Mamba or All Attention Mixers of a Model Are Ablated Together (Real) or Perturbed Together With Size-Matched Shuffled Perturbations (Null), and the Ratio of Each to the Sum of the Corresponding Single-Block Effects. Ten Permutations per Null. Independent Null: Own Permutation per Block; Shared Null: One Permutation per Stage for All Blocks of the Group, Which Keeps the Alignment Between Their Perturbations}
\label{tab:group}
\begin{adjustbox}{max width=\columnwidth}\begin{tabular}{@{}lcccccc@{}}
\toprule
& \multicolumn{3}{c}{all Mamba mixers} & \multicolumn{3}{c}{all attention mixers}\\
\cmidrule(lr){2-4}\cmidrule(l){5-7}
Model & real & null & $\div\Sigma$: real/null & real & null & $\div\Sigma$: real/null\\
\midrule
MV-T, independent & $-$3.90 & $-$1.27 & 3.67 / 1.14 & $-$20.58 & $-$4.22 & 6.96 / 1.36\\
\quad shared & & $-$1.80 & \phantom{3.67 / }1.62 & & $-$7.81 & \phantom{6.96 / }2.52\\
MV-S, independent & $-$3.14 & $-$0.96 & 4.04 / 1.16 & $-$9.31 & $-$2.35 & 5.20 / 1.27\\
\quad shared & & $-$1.54 & \phantom{4.04 / }1.87 & & $-$3.82 & \phantom{5.20 / }2.05\\
MV-B, independent & $-$2.42 & $-$0.40 & 4.76 / 1.60 & $-$9.10 & $-$1.10 & 8.44 / 1.40\\
\quad shared & & $-$0.40 & \phantom{4.76 / }1.59 & & $-$1.10 & \phantom{8.44 / }1.39\\
\bottomrule
\end{tabular}\end{adjustbox}
\end{table}

\noindent\textit{In short:} whole groups of same-type blocks carry far more than their parts, and this survives aligned, size-matched nulls in all three model sizes; for pairs it mostly does not.

If blocks were independent, removing several would cost the sum of removing each. Removing all six Mamba or all six attention mixers of MambaVision-T instead costs $3.67\times$ and $6.96\times$ that sum (\cref{tab:group}), and the ratio for the Mamba group grows with model size ($4.04\times$ in S, $4.76\times$ in B). Group nulls, which apply size-matched shuffled perturbations to the same blocks at once, compound much less. A group null with an independent permutation in every block, however, also destroys the alignment between the perturbations of different blocks, which the real group ablation keeps; aligned perturbations add up in the residual stream, so part of the gap could reflect alignment rather than content. We therefore ran, for all three sizes and with ten permutations each, a second group null that applies one shared permutation to all blocks of a stage, which keeps the pairwise alignment of their perturbations. Alignment matters: the shared null is harsher than the independent one ($-1.80$ vs.\ $-1.27$\pp for the Mamba group, $-7.81$ vs.\ $-4.22$\pp for the attention group) and compounds to $1.62\times$ and $2.52\times$ the sum of single nulls. Yet the real group ablation remains far more damaging. Against the shared null its content effect is $+2.10$\pp $[+1.90,+2.33]$ for the Mamba group and $+12.77$\pp $[+12.45,+13.09]$ for the attention group, about 80\% of the effect measured against the independent null ($+2.63$ and $+16.37$\pp). MambaVision-S behaves the same way ($+1.60$ and $+5.50$\pp against the shared null, $[+1.42,+1.80]$ and $[+5.23,+5.77]$). In MambaVision-B the shared null is no harsher than the independent one (difference $+0.00$\pp in both groups), which suggests that its blocks' perturbations are not aligned in the first place, and the content effects stay at $+2.02$ and $+8.00$\pp. Against aligned nulls, all six groups therefore compound far more under real ablation ($3.67$--$8.44\times$ the sum of single effects, against $1.39$--$2.52\times$). The real ablation also moves the residual stream after the last block further (MambaVision-T: relative displacement 0.37 and 0.47, against 0.25 and 0.29 for the shared null; 0.34--0.40 against 0.15--0.26 in S and B), although each block's perturbation and the alignment between them are matched, so the network amplifies the loss of content more than an equally large and equally aligned perturbation. The biggest type difference in the model, 20.6 versus 3.9\pp, is therefore not a matter of perturbation size or alignment; it is, however, still confounded with depth.

Pairwise ablation localises the compensation (\cref{fig:pairs}a). Of 66 pairs, 9 interact significantly after Holm correction, all negatively and all within one block type: 6 of 15 attention pairs, 3 of 15 Mamba pairs and none of 36 mixed pairs. Seven of the nine are direct neighbours. Adjacency and type are almost always confounded here, but two pairs break the tie: L2B3+L2B4 and L3B1+L3B2 are neighbours of different types, and neither interacts ($-0.10$\pp $[-0.23,+0.04]$ and $-0.11$\pp $[-0.25,+0.04]$). Compensation is therefore between blocks of the same type, not simply between neighbours.

Is the compounding a property of the blocks' content, or would any two perturbations of that size compound? For the four strongest attention pairs we repeated the experiment with size-matched shuffled perturbations of both blocks (\cref{fig:pairs}b). With an independent permutation in each block, the null interactions are close to zero ($-0.29$ to $+0.02$\pp), whereas the real interactions are $-0.49$ to $-2.16$\pp. As for the groups, however, independent permutations destroy the alignment between the two perturbations. With one shared permutation for both blocks (ten permutations), the nulls compound as well: their interactions are $-0.40$ to $-1.32$\pp, and pair over sum of singles is $1.33$--$1.69\times$ against $1.44$--$2.26\times$ for the real pairs. Only the strongest pair, L2B4+L2B5, still compounds significantly more under real ablation ($-0.84$\pp $[-1.09,-0.61]$, Holm $p=0.002$); for the other three the difference is under 0.1\pp and not significant. Most of the pairwise compounding is therefore a property of two aligned perturbations of that size, and only part of it, concentrated in the first two attention mixers of stage 2, is specific to what the blocks compute. Content-specific compensation is clear at the level of whole groups and weak at the level of pairs, so single-block ablation understates what groups of same-type blocks carry, and the natural unit of attribution in this model is a group of same-type blocks rather than a block or a pair.

\section{Selectivity is collective and lives in $B$ and $C$}
\label{sec:selectivity}

\begin{table}[t]
\centering\footnotesize\setlength{\tabcolsep}{3pt}
\caption{Joint freezing of the scan parameters in all six Mamba mixers of MambaVision-T. Share: fraction of the damage of disabling all Mamba mixers ($-$3.90\pp), with a paired bootstrap CI that resamples numerator and denominator together. Mean mode removes image- and position-dependence, position mode image-dependence only.}
\label{tab:freeze}
\resizebox{\linewidth}{!}{\begin{tabular}{@{}llccc@{}}
\toprule
Frozen & mode & $\Delta$ acc.\ (pp) [95\% CI] & share (\%) & log-prob.\ drop\\
\midrule
$\Delta,B,C$ & mean & $-$1.136 [$-$1.328, $-$0.954] & 29.1 [24.8, 33.5] & $-$0.114\\
 & position & $-$0.790 [$-$0.958, $-$0.628] & 20.3 [16.4, 24.3] & $-$0.116\\
\midrule
$\Delta$ & mean & $-$0.540 [$-$0.692, $-$0.390] & 13.8 [10.2, 17.5] & $-$0.039\\
 & position & $-$0.152 [$-$0.274, $-$0.034] & 3.9 [0.9, 6.9] & $-$0.054\\
$B$ & mean & $-$0.274 [$-$0.396, $-$0.146] & 7.0 [3.8, 10.1] & $-$0.079\\
 & position & $-$0.250 [$-$0.370, $-$0.132] & 6.4 [3.5, 9.4] & $-$0.054\\
$C$ & mean & $-$0.220 [$-$0.358, $-$0.082] & 5.6 [2.1, 9.0] & $-$0.098\\
 & position & $-$0.200 [$-$0.326, $-$0.082] & 5.1 [2.1, 8.3] & $-$0.084\\
\bottomrule
\end{tabular}}
\end{table}

\paragraph{Per block, selectivity is nearly invisible.}
Freezing $\Delta$, $B$ and $C$ in a single Mamba mixer of MambaVision-T costs at most 0.23\pp, and freezing $\Delta$ alone to its per-position profile costs at most 0.03\pp in five of six blocks. Read in isolation, these numbers would say that MambaVision barely uses input-dependent selection. The per-block freezes already show the change of confidence discussed below: in the stage-3 mixers L3B0 and L3B1, 16 of the 18 freezing configurations lower accuracy but raise the true-class probability (log-probability drop down to $-0.11$ nats; full grid in \cref{app:grid}).

\begin{figure}[t]
  \centering
  \includegraphics[width=\linewidth]{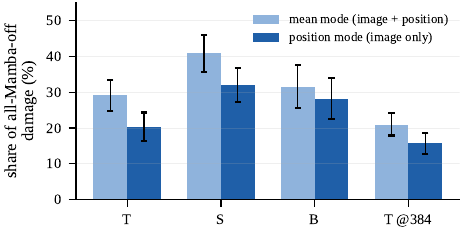}
  \caption{Selectivity fraction across models. Damage of jointly freezing $\Delta,B,C$ in all Mamba mixers as a share of the damage of disabling them, with paired bootstrap 95\% CIs. Position mode isolates image-dependence. T@384: MambaVision-T at 384\,px with global windows (\cref{sec:robust}).}
  \label{fig:selectivity}
\end{figure}

\paragraph{Jointly, it accounts for a fifth to a third of the Mamba contribution.}
Frozen in all six mixers at once, the same intervention costs $1.14$\pp, $1.54\times$ the sum of the per-block freezes (\cref{tab:freeze}). Position mode, which removes only the dependence on the image and keeps a fixed profile over tokens, still costs $0.79$\pp, or 20.3\% $[16.4, 24.3]$ of the damage of disabling all Mamba mixers. The position-mode fraction replicates across model sizes (\cref{fig:selectivity}): 32.1\% $[27.2, 36.9]$ for MambaVision-S and 28.2\% $[22.5, 33.9]$ for MambaVision-B. The tensors play different roles. For $\Delta$, most of the effect is the position profile ($0.54$\pp in mean mode, $0.15$\pp in position mode), whereas $B$ and $C$ cost almost the same in both modes, so their contribution is image-dependence. Selection in MambaVision therefore lives mainly in $B$ and $C$, while the step size behaves close to a fixed, position-dependent schedule. Like the group effects of \cref{sec:compensation}, selectivity is hidden by redundancy when it is tested one block at a time.

\paragraph{The freezing detects selectivity where it is needed.}
A modest selectivity fraction could mean that the model does not need selection, or that the freezing method cannot see it. We applied the same code to mamba-130m on a task that requires selection: random token sequences repeated twice, where the second copy can only be predicted from context~\cite{jelassi2024copying,olsson2022induction}. The intact model copies (\cref{tab:pc}): its loss on the second copy is 5.81 nats lower than on the first. Freezing $B$ or $C$ alone, in either mode, removes almost the entire benefit and sets copy accuracy to zero, while the loss on the first copy stays near its baseline. Freezing only $\Delta$ to its position profile keeps part of the ability (benefit $2.70$ nats). The method thus detects selectivity when a task depends on it, and it shows the same split between $\Delta$ and $B,C$ as in the vision model.

\begin{table}[t]
\centering\footnotesize\setlength{\tabcolsep}{3pt}
\caption{Positive control: the freezing code removes copying in a language Mamba. mamba-130m, 256 sequences of 256 random tokens repeated twice; all 24 layers frozen. Copy benefit: cross-entropy on the first copy minus on the second copy (nats).}
\label{tab:pc}
\begin{tabular}{@{}llcccc@{}}
\toprule
Frozen & mode & copy acc.\,\% & loss 1st & loss 2nd & benefit\\
\midrule
none & -- & 19.06 & 12.75 & 6.94 & $+$5.81\\
$B$ & mean & 0.00 & 12.81 & 12.42 & $+$0.39\\
 & position & 0.00 & 12.71 & 12.47 & $+$0.23\\
$C$ & mean & 0.00 & 12.60 & 12.14 & $+$0.46\\
 & position & 0.00 & 12.59 & 12.22 & $+$0.37\\
$\Delta$ & mean & 0.00 & 13.21 & 12.71 & $+$0.50\\
 & position & 2.46 & 13.40 & 10.70 & $+$2.70\\
$\Delta,B,C$ & mean & 0.00 & 15.11 & 14.37 & $+$0.73\\
 & position & 0.00 & 13.29 & 14.76 & $-$1.48\\
\bottomrule
\end{tabular}
\end{table}

\begin{figure}[t]
  \centering
  \includegraphics[width=\linewidth]{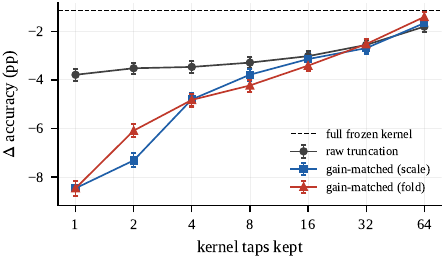}
  \caption{The frozen filter needs its long-lag taps. Accuracy change of MambaVision-T with all six Mamba mixers replaced by their frozen convolution kernels, truncated to the first $k$ taps, with and without restoring the kernel gain (95\% CIs). Dashed: untruncated kernel.}
  \label{fig:truncation}
\end{figure}

\paragraph{The frozen scan is an exact long convolution.}
With selection frozen, each Mamba mixer is a linear time-invariant filter. Implemented explicitly as a causal depthwise convolution, it reproduces the frozen model on 100\% of images. Truncating the kernel shows that the filter needs its long-lag taps (\cref{fig:truncation}): keeping the first 32 taps costs 1.4--1.6\pp more than the full kernel, and even 64 taps leave a gap of 0.27--0.67\pp. Truncation also lowers the kernel's DC gain, so we repeated it with the gain restored, either by rescaling the kept taps or by adding the removed tail to the last kept tap. Neither recovers the loss, and for eight or fewer taps both make it worse, because concentrating the gain in a few taps amplifies token-level variation that the long kernel averages out. This qualifies the observation that Mamba mixers interact locally (\cref{app:range}): most of the weight of the implicit attention map is near the diagonal, yet the small long-range tail is causally needed.

\paragraph{Removing selection changes confidence.}
The joint freeze lowers accuracy but raises the true-class log-probability on average (log-probability drop $-0.11$ in mean and position mode, \cref{tab:freeze}; $-0.68$ and $-0.60$ at 384\,px). Calibration explains this. The unmodified MambaVision-T is strongly under-confident: its mean confidence is 0.61 at 82.2\% accuracy, with an ECE of 21.2\%. This is a property of its training recipe rather than of our pipeline. The released MambaVision models are trained with label smoothing of 0.1, mixup with $\alpha=0.8$ and CutMix with $\alpha=1.0$ (the published training configuration)~\cite{zhang2018mixup,yun2019cutmix}. Label smoothing caps the confidence a network learns to express~\cite{muller2019labelsmoothing}, and mixup trains on soft targets of mixed images; combined, they are known to leave networks under-confident on clean test images~\cite{wang2023pitfall,thulasidasan2019mixup}. With $\Delta$, $B$ and $C$ frozen in all Mamba mixers, mean confidence rises to 0.73 (mean mode) and 0.72 (position mode), predictive entropy falls from 2.90 to 1.88--1.98 nats, the negative log-likelihood falls from 0.968 to 0.852--0.854, and ECE drops to 8.5--9.8\%, while accuracy falls by 1.1 and 0.8\pp. Confidence rises on wrong predictions as well (0.40 to 0.48). The lower ECE should therefore not be read as better calibration in any useful sense. Confidence moves towards accuracy because the frozen model loses some of the softening it learned from smoothed and mixed targets, not because its probabilities become more trustworthy: it commits more strongly, including on images it gets wrong. Input-dependent selection thus contributes to the soft predictions this recipe produces, and the change of confidence is one reason why accuracy and log-probability can disagree in sign.

\section{Reference Distribution and Distribution Shift}
\label{sec:robust}

The findings above use clean $224\times224$ images and reference statistics from the same distribution. We test both choices.

\paragraph{Reference distribution.} Mean ablation replaces activations by a reference mean, and the reference matters. At 384\,px (with attention windows enlarged to the full token grid, a setting the model was not trained on), the per-block content effects on the first 10,000 images, pooled over four permutations, are $+0.23$\pp $[+0.13,+0.33]$ when the reference means are recomputed at 384\,px, but $-0.005$\pp $[-0.10,+0.09]$ when the 224\,px means are reused. A mismatched reference adds an off-distribution component to both the real and the null perturbation and hides the difference between them. At 384\,px, the selectivity fraction falls to 15.6\% (\cref{fig:selectivity}) although the Mamba mixers matter more ($-5.9$\pp when all are disabled), so in this one setting, which lies outside the model's training regime, longer token sequences did not make selection more important.

\paragraph{Distribution shift.} Under four ImageNet-C corruptions~\cite{hendrycks2019imagenetc} at severity 3 (first 10,000 images, ten null permutations per block, clean-image reference means), real and null damage separate in more blocks than on clean images: 0, 4, 4 and 5 of 12 blocks after correction for contrast, JPEG compression, Gaussian noise and defocus blur. The separations go both ways. Under JPEG compression and defocus blur most significant blocks are null-worse (L2B0, L2B4 and L2B5), and only under Gaussian noise does the real ablation dominate, with the early Mamba mixers L2B0, L2B2 and L2B3 losing up to 4.1\pp more than their nulls. The median ratio $|C|/|\text{null}|$ over blocks is 0.25 on clean images and 0.26--0.58 under corruption, largest for Gaussian noise, and the pooled effect changes sign between conditions ($-0.21$ to $+0.24$\pp). Content-specific effects therefore do appear under distribution shift, most clearly for the early Mamba mixers under noise, but they are neither uniform nor of one sign. Per-block results are in \cref{app:corrupt}.

\section{Discussion}
\label{sec:discussion}

\subsection{What the Results Mean for Interpreting MambaVision}
A single-block ablation score in MambaVision is small, and to a close approximation it is a map of where the residual stream is sensitive to a perturbation of a given size. Its content-specific part is small in absolute terms, but relative to the damage it is of the same order as in a hierarchical Transformer (\cref{sec:null}). Read as a ranking of block importance, it would say that the stage-2 attention mixers matter most and the Mamba mixers least. Our nulls show that this ranking is reproduced by perturbations that carry no information about the blocks, so it does not license claims about what the blocks compute. The content of the blocks shows up at a different scale. Groups of same-type blocks carry several times more than their parts, and the compounding is much weaker when the removed content is replaced by a size-matched null, even one that keeps the alignment between the blocks' perturbations. For adjacent pairs, by contrast, aligned nulls reproduce most of the compounding. The natural unit of attribution in this model is therefore a group of neighbouring blocks of one type, not a block.

The selectivity results bear on the debate about whether vision needs Mamba~\cite{yu2025mambaout,han2024demystify}. In a trained MambaVision the Mamba mixers do contribute (3.9\pp of accuracy for the six mixers of MambaVision-T), but most of this contribution survives when selection is frozen and the scan becomes a fixed long convolution. The part that needs input-dependent selection, 20--32\% depending on model size, is carried by $B$ and $C$, which play the roles of keys and queries in the linear-attention reading of the scan~\cite{han2024demystify}, whereas the step size acts close to a fixed, position-dependent schedule. These findings are compatible with both views: selection is used, but it is not where most of the Mamba contribution comes from.

The metric matters as much as the intervention. In accuracy and true-class rank, single blocks show almost no content effect; in log-probability they show small, consistent ones. The rank results identify these as changes of confidence. Interventions on MambaVision often change how sharply the model commits without changing which classes it ranks highest, and the model's training recipe makes this especially visible (\cref{sec:selectivity}).

\subsection{Limitations}
\paragraph{Type and depth.} In MambaVision every Mamba mixer precedes every attention mixer within a stage, so each type comparison is also a depth comparison. Our nulls separate content from magnitude and position, not mixer type from depth. Separating the two needs hybrids whose block placement varies, for example models trained with randomised placement.

\paragraph{One model family and one control.} We study three sizes of one hybrid family, and we restrict our claims to MambaVision. DeiT-Small and Swin-T show that the test detects content effects in flat and in hierarchical attention-only models. Relative to damage, however, MambaVision's content share is comparable to Swin-T's in accuracy, so what distinguishes MambaVision in our data is mainly the small size of its single-block damage, not an absence of content effects. Which feature produces this robustness, the Mamba mixers themselves, their placement before attention within each stage, the convolutional early stages or the training recipe, cannot be separated with controls of this kind; a second hybrid family is needed.

\paragraph{Permutations.} Every per-block shuffled null uses ten permutations, except the 384-pixel experiment of \cref{sec:robust}, which uses four. A single permutation can mislead in individual blocks (L2B6, L2B7 and L3B0 of MambaVision-B; S0B0 of Swin-T), whereas pooled conclusions were stable; the pair nulls were tested with aligned perturbations only in MambaVision-T; there, keeping that alignment removed most of the pairwise content effects, while it reduced the group content effects by about a fifth in T and S and not at all in B.

\paragraph{Data and seeds.} All results are on ImageNet-1k classification. Only one checkpoint per model size has been released, so we cannot measure how much the effects vary between training runs; we anchor the equivalence margin to the spread of the null instead (\cref{sec:stats}). The reference images are part of the evaluation set, which changes no effect by more than 0.024\pp.

\paragraph{What the null controls.} The shuffled null matches the size and the location of the real perturbation, not its direction. $C$ is therefore the part of the damage that depends on the direction of the removed signal, given its size and where it enters, and directions to which the downstream network is more sensitive will show up as content. Two checks limit how much of $C$ can be a generic sensitivity to high-variance channels: the sign-flip null keeps the magnitude of every channel in place and only changes signs, and the resample null replaces the output by an in-distribution sample. Both lead to the same conclusions as the shuffled null in the per-block tests (\cref{tab:metric} and \cref{sec:null}).

\paragraph{Nulls.} The shuffled and sign-flip nulls are synthetic perturbations of the real ablation. Resample ablation is an in-distribution check, but it is not size-matched: its perturbation is larger than that of mean ablation, so it tests the ranking of blocks rather than the equivalence bound.

\subsection{Future Work}
Four directions follow. First, hybrids with varied block placement would separate type from depth. Second, the protocol applies unchanged to other hybrid families and to dense prediction tasks, where the Mamba mixers may matter more. Third, circuit-discovery methods that prune components by their ablation effect, such as ACDC~\cite{conmy2023acdc}, could compare each effect with a size-matched null before pruning. Fourth, the observation that freezing selection sharpens predictions suggests studying how input-dependent selection interacts with soft-target training more generally.

\section{A Protocol for Causal Attribution in Hybrid Networks}
\label{sec:checklist}

\Cref{tab:checklist} turns the results into a protocol. Each step addresses one way in which block ablation misled us, and the right-hand column follows one block through all of them. L2B4, the first attention mixer of stage 2 in MambaVision-T, is the natural test case: its ablation costs more accuracy than that of any other block, so a standard analysis would call it the most important block in the model.

\begin{table*}[t]
\centering
\caption{Checklist for Block-Level Causal Attribution, With a Worked Example on Block L2B4 of MambaVision-T (50,000 Images)}
\label{tab:checklist}
\footnotesize
\renewcommand{\arraystretch}{1.25}
\begin{tabular}{@{}p{0.03\textwidth}p{0.40\textwidth}p{0.52\textwidth}@{}}
\toprule
& Step & Worked example: L2B4\\
\midrule
1 & \textit{Build the reference set from the input distribution under test} and record how it was built. Recompute it when the resolution or input distribution changes, and re-score effects on images outside it. & Means over the first 256 images of the fixed order. Excluding all reference images changes no effect by more than 0.024\pp. At 384\,px, reusing 224\,px means would have hidden a pooled content effect of $+0.23$\pp. \\
2 & \textit{Ablate, and test the effect against the unmodified model} with a paired test on the same images. & Mean ablation costs 1.13\pp (McNemar $p<10^{-38}$), the largest single-block drop in the model. \\
3 & \textit{Run a size-matched null with several permutations, and a second kind of null,} on the same images. Report $C$ with a paired bootstrap interval and correct over blocks. & Shuffled null (ten permutations): $-1.29\pm0.08$\pp, so $C=-0.16$\pp $[-0.30,-0.00]$. Sign-flip null: $-1.35$\pp, $C=-0.21$\pp $[-0.38,-0.04]$. Neither survives Holm correction. The null damages L2B4 at least as much as its real ablation. \\
4 & \textit{Relate damage to perturbation size} to see whether a position is fragile. & Across the attention positions, null damage grows by 5.4\pp per unit of relative perturbation size ($r=1.00$). L2B4 receives the largest perturbation of all blocks (0.30 of the residual norm) and shows the largest damage. Its ranking reflects position, not content. \\
5 & \textit{Pool over blocks and test equivalence} against a margin fixed in advance and anchored to the null's own spread, \textit{and scale content effects by damage}, so that small damage is not mistaken for absent content. & Pooled over the 12 blocks, the 90\% interval $[-0.044,+0.016]$\pp lies inside $\pm0.10$\pp; the null of L2B4 alone varies by 0.20\pp across permutations. Relative to damage, the content share of the model is $R=0.20$ $[0.06,0.31]$, comparable to Swin-T's. \\
6 & \textit{Report accuracy, log-probability and true-class rank}, and read disagreement between them as a change of confidence. & Log-probability: the null lowers $p(y)$ more ($C=-0.030$ and $-0.037$ nats). Rank: real 0.0238, nulls 0.0235 and 0.0229. The ordering of classes changes equally under real and null. \\
7 & \textit{Intervene on groups and adjacent pairs, with group nulls} that keep the alignment between the blocks' perturbations, and compare with the sum of single effects. & Ablating all six attention mixers costs 20.58\pp, $6.96\times$ the sum of their single effects; group nulls cost 4.22\pp with independent and 7.81\pp with shared permutations ($1.36\times$ and $2.52\times$). Of L2B4's pairs only L2B4+L2B5 compounds more than its aligned null ($-0.84$\pp). L2B4's content is carried jointly with the other attention mixers. \\
8 & \textit{Test a mechanism by freezing it jointly}, separating image- from position-dependence, and validate the freezing with a positive control. & (Applies to Mamba mixers.) Joint freezing of $\Delta$, $B$, $C$ in all six Mamba mixers costs 20.3\% of their contribution in position mode; the same code removes in-context copying in mamba-130m. \\
\bottomrule
\end{tabular}
\end{table*}

Followed to the end, the protocol reverses the standard reading of L2B4. The block sits at the most fragile position in the residual stream, its single-block ablation is reproduced by perturbations that carry none of its content, and what it computes becomes visible only when it is removed together with the other attention mixers. None of the individual steps is expensive: the nulls cost one extra forward pass per permutation, and several nulls can share the forward pass up to the ablated block.

\section{Conclusion}
Block ablation in MambaVision mostly measures where the residual stream is fragile. Once the size and location of the perturbation are controlled with size-matched nulls, little content-specific damage remains per block in accuracy and in the rank of the true class; relative to its small damage, this share is comparable to that of a hierarchical Transformer in accuracy and lower in rank, while groups of same-type blocks carry far more than their parts, beyond what equally large and equally aligned perturbations explain. Input-dependent selection is real but collective, sits mainly in $B$ and $C$, and its frozen form is an exact long convolution. The protocol of \cref{sec:checklist} makes these controls routine.

\appendices
\crefalias{section}{appendix}
\section{Per-Block Results}
\label{app:perblock}
\Cref{tab:a_T} lists every single-block result for MambaVision-T and \cref{tab:a_S,tab:a_B,tab:a_deit,tab:a_swin} the corresponding results for MambaVision-S, MambaVision-B, DeiT-Small and Swin-T. $C$ is the content effect of \cref{sec:stats}; Holm-corrected $p$ values are over the blocks of each model.

\begin{table}[!t]
\centering
\caption{MambaVision-T: Single-Block Mean Ablation and the Ten-Permutation Shuffled Null (50,000 Images)}
\label{tab:a_T}
\footnotesize\setlength{\tabcolsep}{2.2pt}
\begin{adjustbox}{max width=\columnwidth}\begin{tabular}{@{}llcccc@{}}
\toprule
Block & type & real (pp) & null (pp), mean $\pm$ SD & $C$ (pp) [95\% CI] & Holm $p$\\
\midrule
L2B0 & M & $-$0.282 & $-$0.201 $\pm$ 0.028 & $+$0.081 [$-$0.024, $+$0.186] & 0.8689\\
L2B1 & M & $-$0.084 & $-$0.107 $\pm$ 0.048 & $-$0.023 [$-$0.111, $+$0.068] & 1.0000\\
L2B2 & M & $-$0.102 & $-$0.109 $\pm$ 0.061 & $-$0.007 [$-$0.090, $+$0.071] & 1.0000\\
L2B3 & M & $-$0.124 & $-$0.165 $\pm$ 0.064 & $-$0.041 [$-$0.122, $+$0.041] & 1.0000\\
L2B4 & A & $-$1.132 & $-$1.295 $\pm$ 0.080 & $-$0.163 [$-$0.305, $-$0.003] & 0.3125\\
L2B5 & A & $-$0.582 & $-$0.618 $\pm$ 0.070 & $-$0.036 [$-$0.161, $+$0.089] & 1.0000\\
L2B6 & A & $-$0.396 & $-$0.528 $\pm$ 0.050 & $-$0.132 [$-$0.246, $-$0.012] & 0.3042\\
L2B7 & A & $-$0.288 & $-$0.407 $\pm$ 0.057 & $-$0.119 [$-$0.237, $-$0.005] & 0.3125\\
L3B0 & M & $-$0.262 & $-$0.250 $\pm$ 0.040 & $+$0.012 [$-$0.094, $+$0.117] & 1.0000\\
L3B1 & M & $-$0.208 & $-$0.280 $\pm$ 0.048 & $-$0.072 [$-$0.185, $+$0.041] & 1.0000\\
L3B2 & A & $-$0.328 & $-$0.204 $\pm$ 0.065 & $+$0.124 [$+$0.007, $+$0.234] & 0.3125\\
L3B3 & A & $-$0.232 & $-$0.042 $\pm$ 0.045 & $+$0.190 [$+$0.097, $+$0.285] & 0.0008\\
\bottomrule
\end{tabular}\end{adjustbox}
\end{table}

\begin{table}[!t]
\centering
\caption{MambaVision-S: Real Ablation Versus Ten Shuffled-Null Permutations per Block (50,000 Images). $^{*}$ Holm $p<0.05$ Over 12 Blocks; Log-Prob.\ in Nats, Rank in $\log(1+\text{rank})$}
\label{tab:a_S}
\footnotesize\setlength{\tabcolsep}{2.2pt}
\begin{adjustbox}{max width=\columnwidth}\begin{tabular}{@{}llcccccc@{}}
\toprule
Block & type & real (pp) & null (pp) & null SD & $C$ acc.\ (pp) [95\% CI] & $C$ log-prob. & $C$ rank\\
\midrule
L2B0 & M & $-$0.178 & $-$0.149 & 0.047 & $+$0.029 [$-$0.066, $+$0.122] & $+$0.016$^{*}$ & $+$0.001\\
L2B1 & M & $-$0.066 & $-$0.063 & 0.028 & $+$0.003 [$-$0.082, $+$0.083] & $+$0.024$^{*}$ & $+$0.001\\
L2B2 & M & $-$0.028 & $-$0.052 & 0.031 & $-$0.024 [$-$0.103, $+$0.055] & $+$0.021$^{*}$ & 0.000\\
L2B3 & M & $-$0.134 & $-$0.036 & 0.055 & $+$0.098 [$+$0.020, $+$0.178] & $+$0.017$^{*}$ & $+$0.002$^{*}$\\
L2B4 & A & $-$0.746 & $-$1.071 & 0.104 & $-$0.325$^{*}$ [$-$0.460, $-$0.191] & $-$0.010$^{*}$ & $-$0.001\\
L2B5 & A & $-$0.318 & $-$0.368 & 0.063 & $-$0.050 [$-$0.153, $+$0.063] & $-$0.005$^{*}$ & $+$0.001\\
L2B6 & A & $-$0.300 & $-$0.300 & 0.079 & 0.000 [$-$0.109, $+$0.114] & $-$0.003$^{*}$ & $+$0.001\\
L3B0 & M & $-$0.182 & $-$0.161 & 0.035 & $+$0.021 [$-$0.084, $+$0.129] & $+$0.018$^{*}$ & 0.000\\
L3B1 & M & $-$0.174 & $-$0.211 & 0.073 & $-$0.037 [$-$0.144, $+$0.070] & $+$0.012$^{*}$ & $-$0.001\\
L3B2 & M & $-$0.016 & $-$0.154 & 0.051 & $-$0.138 [$-$0.238, $-$0.040] & $-$0.123$^{*}$ & $-$0.002$^{*}$\\
L3B3 & A & $-$0.348 & $-$0.103 & 0.040 & $+$0.245$^{*}$ [$+$0.147, $+$0.344] & $-$0.088$^{*}$ & $-$0.001\\
L3B4 & A & $-$0.080 & $-$0.016 & 0.020 & $+$0.064 [$-$0.017, $+$0.146] & $-$0.055$^{*}$ & $-$0.001\\
\bottomrule
\end{tabular}\end{adjustbox}
\end{table}

\begin{table}[!t]
\centering
\caption{MambaVision-B: Real Ablation Versus Ten Shuffled-Null Permutations per Block, Size-Matched in the Residual Stream (50,000 Images). $^{*}$ Holm $p<0.05$ Over 15 Blocks; Log-Prob.\ in Nats, Rank in $\log(1+\text{rank})$}
\label{tab:a_B}
\footnotesize\setlength{\tabcolsep}{2.2pt}
\begin{adjustbox}{max width=\columnwidth}\begin{tabular}{@{}llcccccc@{}}
\toprule
Block & type & real (pp) & null (pp) & null SD & $C$ acc.\ (pp) [95\% CI] & $C$ log-prob. & $C$ rank\\
\midrule
L2B0 & M & $-$0.118 & $-$0.064 & 0.055 & $+$0.054 [$-$0.028, $+$0.135] & $+$0.001 & $+$0.001\\
L2B1 & M & $-$0.036 & $-$0.029 & 0.035 & $+$0.007 [$-$0.069, $+$0.082] & $+$0.022$^{*}$ & $+$0.001$^{*}$\\
L2B2 & M & $-$0.050 & $-$0.028 & 0.035 & $+$0.022 [$-$0.038, $+$0.085] & $+$0.017$^{*}$ & $+$0.001\\
L2B3 & M & $-$0.016 & $-$0.027 & 0.026 & $-$0.011 [$-$0.073, $+$0.056] & $+$0.018$^{*}$ & 0.000\\
L2B4 & M & $-$0.022 & $-$0.019 & 0.015 & $+$0.003 [$-$0.059, $+$0.069] & $+$0.026$^{*}$ & $+$0.001\\
L2B5 & A & $-$0.320 & $-$0.268 & 0.027 & $+$0.052 [$-$0.047, $+$0.154] & $+$0.012$^{*}$ & $+$0.002$^{*}$\\
L2B6 & A & $-$0.170 & $-$0.138 & 0.043 & $+$0.032 [$-$0.058, $+$0.122] & $+$0.013$^{*}$ & $+$0.003$^{*}$\\
L2B7 & A & $-$0.136 & $-$0.134 & 0.063 & $+$0.002 [$-$0.087, $+$0.094] & $+$0.004$^{*}$ & $+$0.001\\
L2B8 & A & $-$0.122 & $-$0.124 & 0.045 & $-$0.002 [$-$0.091, $+$0.098] & $+$0.007$^{*}$ & $+$0.001\\
L2B9 & A & $-$0.134 & $-$0.109 & 0.041 & $+$0.025 [$-$0.063, $+$0.111] & $-$0.007$^{*}$ & $+$0.001\\
L3B0 & M & $-$0.142 & $-$0.056 & 0.032 & $+$0.086 [$-$0.015, $+$0.184] & $+$0.007$^{*}$ & $-$0.001\\
L3B1 & M & $-$0.050 & $-$0.021 & 0.027 & $+$0.029 [$-$0.059, $+$0.117] & $-$0.025$^{*}$ & $-$0.001\\
L3B2 & M & $-$0.074 & $-$0.006 & 0.022 & $+$0.068 [$-$0.014, $+$0.149] & $-$0.046$^{*}$ & $-$0.001\\
L3B3 & A & $-$0.140 & $-$0.005 & 0.027 & $+$0.135$^{*}$ [$+$0.055, $+$0.218] & $-$0.064$^{*}$ & 0.000\\
L3B4 & A & $-$0.056 & $-$0.007 & 0.019 & $+$0.049 [$-$0.009, $+$0.107] & $-$0.020$^{*}$ & $-$0.001$^{*}$\\
\bottomrule
\end{tabular}\end{adjustbox}
\end{table}

\begin{table}[!t]
\centering
\caption{DeiT-Small: Mean Ablation of Each Attention Layer Versus Ten Shuffled-Null Permutations (50,000 Images). $^{*}$ Holm $p<0.05$ Over 12 Layers; Log-Prob.\ in Nats, Rank in $\log(1+\text{rank})$}
\label{tab:a_deit}
\footnotesize\setlength{\tabcolsep}{2.2pt}
\begin{adjustbox}{max width=\columnwidth}\begin{tabular}{@{}lcccccc@{}}
\toprule
Layer & real (pp) & null (pp) & null SD & $C$ acc.\ (pp) [95\% CI] & $C$ log-prob. & $C$ rank\\
\midrule
L0 & $-$1.356 & $-$10.105 & 0.360 & $-$8.749$^{*}$ [$-$9.023, $-$8.498] & $-$0.471$^{*}$ & $-$0.215$^{*}$\\
L1 & $-$9.658 & $-$5.436 & 0.292 & $+$4.222$^{*}$ [$+$3.961, $+$4.490] & $+$0.221$^{*}$ & $+$0.110$^{*}$\\
L2 & $-$3.926 & $-$1.551 & 0.123 & $+$2.375$^{*}$ [$+$2.168, $+$2.597] & $+$0.104$^{*}$ & $+$0.057$^{*}$\\
L3 & $-$2.428 & $-$1.133 & 0.095 & $+$1.295$^{*}$ [$+$1.102, $+$1.499] & $+$0.057$^{*}$ & $+$0.033$^{*}$\\
L4 & $-$1.212 & $-$0.968 & 0.067 & $+$0.244$^{*}$ [$+$0.083, $+$0.408] & $-$0.008$^{*}$ & $+$0.004$^{*}$\\
L5 & $-$1.102 & $-$1.123 & 0.099 & $-$0.021 [$-$0.177, $+$0.138] & $+$0.004 & $+$0.002\\
L6 & $-$1.472 & $-$1.347 & 0.077 & $+$0.125 [$-$0.048, $+$0.296] & $-$0.005$^{*}$ & $+$0.001\\
L7 & $-$0.986 & $-$1.177 & 0.082 & $-$0.191 [$-$0.353, $-$0.028] & $-$0.013$^{*}$ & $-$0.006$^{*}$\\
L8 & $-$0.910 & $-$0.988 & 0.071 & $-$0.078 [$-$0.240, $+$0.083] & $-$0.013$^{*}$ & $-$0.002\\
L9 & $-$0.800 & $-$0.696 & 0.069 & $+$0.104 [$-$0.045, $+$0.258] & $-$0.019$^{*}$ & 0.000\\
L10 & $-$0.762 & $-$0.414 & 0.072 & $+$0.348$^{*}$ [$+$0.206, $+$0.488] & $-$0.028$^{*}$ & $+$0.004$^{*}$\\
L11 & $-$0.602 & $-$1.578 & 0.106 & $-$0.976$^{*}$ [$-$1.118, $-$0.837] & $+$0.001 & $-$0.055$^{*}$\\
\bottomrule
\end{tabular}\end{adjustbox}
\end{table}

\begin{table}[!t]
\centering
\caption{Swin-T: Mean Ablation of Each Attention Block Versus Ten Shuffled-Null Permutations (50,000 Images). $^{*}$ Holm $p<0.05$ Over 12 Blocks; Log-Prob.\ in Nats, Rank in $\log(1+\text{rank})$}
\label{tab:a_swin}
\footnotesize\setlength{\tabcolsep}{2.2pt}
\begin{adjustbox}{max width=\columnwidth}\begin{tabular}{@{}lcccccc@{}}
\toprule
Block & real (pp) & null (pp) & null SD & $C$ acc.\ (pp) [95\% CI] & $C$ log-prob. & $C$ rank\\
\midrule
S0B0 & $-$6.654 & $-$6.334 & 1.322 & $+$0.320$^{*}$ [$+$0.076, $+$0.572] & $+$0.088$^{*}$ & $+$0.011$^{*}$\\
S0B1 & $-$1.696 & $-$2.351 & 0.181 & $-$0.655$^{*}$ [$-$0.824, $-$0.487] & $-$0.057$^{*}$ & $-$0.016$^{*}$\\
S1B0 & $-$1.848 & $-$1.358 & 0.091 & $+$0.490$^{*}$ [$+$0.311, $+$0.655] & $+$0.038$^{*}$ & $+$0.013$^{*}$\\
S1B1 & $-$1.924 & $-$0.917 & 0.083 & $+$1.007$^{*}$ [$+$0.831, $+$1.189] & $+$0.035$^{*}$ & $+$0.022$^{*}$\\
S2B0 & $-$1.534 & $-$1.232 & 0.129 & $+$0.302$^{*}$ [$+$0.144, $+$0.470] & $+$0.001 & $+$0.010$^{*}$\\
S2B1 & $-$0.884 & $-$0.671 & 0.069 & $+$0.213$^{*}$ [$+$0.067, $+$0.355] & $-$0.009$^{*}$ & $+$0.007$^{*}$\\
S2B2 & $-$0.996 & $-$0.687 & 0.079 & $+$0.309$^{*}$ [$+$0.155, $+$0.456] & $+$0.005$^{*}$ & $+$0.011$^{*}$\\
S2B3 & $-$0.446 & $-$0.466 & 0.070 & $-$0.020 [$-$0.133, $+$0.103] & $-$0.004$^{*}$ & $+$0.002$^{*}$\\
S2B4 & $-$0.912 & $-$0.778 & 0.060 & $+$0.134 [$-$0.004, $+$0.268] & $-$0.006$^{*}$ & $+$0.002$^{*}$\\
S2B5 & $-$0.478 & $-$0.587 & 0.051 & $-$0.109 [$-$0.228, $+$0.011] & $-$0.001 & 0.000\\
S3B0 & $-$1.750 & $-$1.637 & 0.084 & $+$0.113 [$-$0.065, $+$0.281] & $-$0.094$^{*}$ & $-$0.005$^{*}$\\
S3B1 & $-$0.806 & $-$2.056 & 0.180 & $-$1.250$^{*}$ [$-$1.398, $-$1.102] & $-$0.100$^{*}$ & $-$0.051$^{*}$\\
\bottomrule
\end{tabular}\end{adjustbox}
\end{table}

\section{Channel Statistics and Rank Damage}
\label{app:channels}
\Cref{tab:a_rms} reports how concentrated the mixer output and the ablation perturbation are in a few channels, measured on the 256 reference images. The perturbation, not the output, determines how much mass the shuffled null moves between channels. \Cref{tab:a_rank} gives the damage in top-1 accuracy, true-class rank and top-5 accuracy for the real ablation and both nulls (mean over three permutations or seeds); the content effects in rank discussed in \cref{sec:null} are the differences between these columns.

\begin{table}[!t]
\centering
\caption{Concentration of the Mixer Output and of the Ablation Perturbation $\Delta=\bar{x}-x$ in Few Channels (MambaVision-T, 256 Reference Images)}
\label{tab:a_rms}
\footnotesize\setlength{\tabcolsep}{2.2pt}
\begin{adjustbox}{max width=\columnwidth}\begin{tabular}{@{}llcccc@{}}
\toprule
& & \multicolumn{2}{c}{mixer output} & \multicolumn{2}{c}{perturbation}\\ \cmidrule(lr){3-4}\cmidrule(l){5-6} Block & type & max/med. & top-1\% & max/med. & top-1\%\\
\midrule
L2B0 & M & 1.88 & 3.0\% & 2.04 & 3.0\%\\
L2B1 & M & 1.89 & 2.5\% & 1.48 & 1.8\%\\
L2B2 & M & 2.11 & 2.6\% & 1.33 & 1.6\%\\
L2B3 & M & 7.80 & 16.5\% & 7.38 & 14.9\%\\
L2B4 & A & 9.43 & 20.0\% & 1.96 & 2.6\%\\
L2B5 & A & 8.69 & 18.6\% & 1.73 & 1.9\%\\
L2B6 & A & 8.47 & 18.3\% & 2.20 & 2.5\%\\
L2B7 & A & 5.14 & 8.6\% & 2.01 & 2.2\%\\
L3B0 & M & 8.86 & 17.6\% & 6.88 & 11.5\%\\
L3B1 & M & 13.65 & 32.2\% & 12.51 & 29.1\%\\
L3B2 & A & 14.83 & 34.8\% & 9.72 & 19.1\%\\
L3B3 & A & 9.53 & 21.6\% & 8.40 & 17.9\%\\
\bottomrule
\end{tabular}\end{adjustbox}
\end{table}

\begin{table}[!t]
\centering
\caption{Damage Under the Real Ablation and Both Nulls in Three Metrics (MambaVision-T, 50,000 Images). Rank: Increase of the Mean of $\log(1+\text{rank})$ Over Its Baseline of 0.232}
\label{tab:a_rank}
\footnotesize\setlength{\tabcolsep}{2.2pt}
\begin{adjustbox}{max width=\columnwidth}\begin{tabular}{@{}lccccccccc@{}}
\toprule
& \multicolumn{3}{c}{$\Delta$ top-1 (pp)} & \multicolumn{3}{c}{rank damage} & \multicolumn{3}{c}{$\Delta$ top-5 (pp)}\\ \cmidrule(lr){2-4}\cmidrule(lr){5-7}\cmidrule(l){8-10} Block & real & shuf. & sign & real & shuf. & sign & real & shuf. & sign\\
\midrule
L2B0 & $-$0.28 & $-$0.19 & $-$0.24 & 0.0054 & 0.0042 & 0.0041 & $-$0.11 & $-$0.13 & $-$0.12\\
L2B1 & $-$0.08 & $-$0.14 & $-$0.14 & 0.0031 & 0.0028 & 0.0021 & $-$0.09 & $-$0.09 & $-$0.03\\
L2B2 & $-$0.10 & $-$0.15 & $-$0.06 & 0.0029 & 0.0027 & 0.0017 & $-$0.00 & $-$0.04 & $-$0.05\\
L2B3 & $-$0.12 & $-$0.15 & $-$0.15 & 0.0032 & 0.0028 & 0.0030 & $-$0.09 & $-$0.05 & $-$0.04\\
L2B4 & $-$1.13 & $-$1.38 & $-$1.34 & 0.0238 & 0.0235 & 0.0229 & $-$0.56 & $-$0.52 & $-$0.54\\
L2B5 & $-$0.58 & $-$0.64 & $-$0.59 & 0.0115 & 0.0115 & 0.0108 & $-$0.23 & $-$0.29 & $-$0.24\\
L2B6 & $-$0.40 & $-$0.49 & $-$0.56 & 0.0085 & 0.0082 & 0.0089 & $-$0.24 & $-$0.19 & $-$0.21\\
L2B7 & $-$0.29 & $-$0.47 & $-$0.39 & 0.0067 & 0.0068 & 0.0062 & $-$0.18 & $-$0.13 & $-$0.14\\
L3B0 & $-$0.26 & $-$0.29 & $-$0.24 & 0.0037 & 0.0043 & 0.0036 & $-$0.06 & $-$0.08 & $-$0.09\\
L3B1 & $-$0.21 & $-$0.24 & $-$0.22 & 0.0026 & 0.0040 & 0.0030 & $-$0.02 & $-$0.10 & $-$0.07\\
L3B2 & $-$0.33 & $-$0.21 & $-$0.13 & 0.0035 & 0.0055 & 0.0036 & $-$0.10 & $-$0.11 & $-$0.08\\
L3B3 & $-$0.23 & $-$0.09 & $-$0.01 & 0.0019 & 0.0022 & 0.0013 & $-$0.11 & $-$0.03 & $-$0.01\\
\bottomrule
\end{tabular}\end{adjustbox}
\end{table}

\section{SSM Pathway Versus Gate Pathway}
\label{app:pathway}
Each Mamba mixer of MambaVision combines a selective-scan branch with output $y$ and a gate branch with output $z$. We replace either output by its per-channel mean over the first 2,000 images and compare the two on the same images (\cref{tab:a_pw}). In the first three Mamba mixers of stage 2, removing the scan branch costs more than removing the gate branch (by 0.11--0.15\pp, 95\% intervals excluding zero); in L2B3 and in stage 3 the two cannot be distinguished. All pathway effects are small (at most 0.22\pp) and have not been compared with size-matched nulls, so they describe sensitivity rather than content.

\begin{table}[!t]
\centering
\caption{Pathway Ablation in the Mamba Mixers of MambaVision-T (50,000 Images). Contrast: Per-Image Difference SSM-Off Minus Gate-Off (Negative: Removing the Scan Branch Is More Damaging)}
\label{tab:a_pw}
\footnotesize\setlength{\tabcolsep}{2.2pt}
\begin{adjustbox}{max width=\columnwidth}\begin{tabular}{@{}llccc@{}}
\toprule
Block & removed & $\Delta$ acc.\ (pp) & log-prob.\ drop & contrast (pp) [95\% CI]\\
\midrule
L2B0 & gate $z$ & $-$0.076 & $+$0.0084 & \\
L2B0 & scan $y$ & $-$0.216 & $+$0.0368 & $-$0.140 [$-$0.252, $-$0.022]\\
L2B1 & gate $z$ & 0.036 & $+$0.0042 & \\
L2B1 & scan $y$ & $-$0.110 & $+$0.0162 & $-$0.146 [$-$0.248, $-$0.044]\\
L2B2 & gate $z$ & $-$0.018 & $+$0.0034 & \\
L2B2 & scan $y$ & $-$0.126 & $+$0.0175 & $-$0.108 [$-$0.216, $-$0.004]\\
L2B3 & gate $z$ & $-$0.038 & $-$0.0013 & \\
L2B3 & scan $y$ & $-$0.094 & $+$0.0212 & $-$0.056 [$-$0.168, $+$0.050]\\
L3B0 & gate $z$ & $-$0.210 & $+$0.0007 & \\
L3B0 & scan $y$ & $-$0.132 & $+$0.0242 & 0.078 [$-$0.060, $+$0.220]\\
L3B1 & gate $z$ & $-$0.176 & $-$0.0660 & \\
L3B1 & scan $y$ & $-$0.140 & $+$0.0983 & 0.036 [$-$0.100, $+$0.168]\\
\bottomrule
\end{tabular}\end{adjustbox}
\end{table}

\section{Sampling Audit}
\label{app:audit}
Reference statistics come from the first 256 evaluation images (pathway analysis: 2,000). Re-scoring 26 effects, covering single blocks, pathway ablations and the two group ablations, on the 48,000 images never used as references changes no effect by more than 0.024\pp. The cached-image evaluation path reproduces all 50,000 baseline predictions of the standard data loader.

\section{Full Freezing Grid}
\label{app:grid}
\Cref{tab:a_grid} lists all 54 single-block freezing configurations of MambaVision-T that \cref{sec:selectivity} summarizes.

\begin{table*}[!t]
\centering
\caption{All 54 Single-Block Freezing Configurations of MambaVision-T (50,000 Images). $\Delta$ acc.: Accuracy Change (pp), $^{\dagger}$ McNemar $p<0.05/54$. Log-p.\ Drop: Mean Decrease of the True-Class Log-Probability (Nats; Negative: Frozen Model Assigns the True Class a Higher Probability), 95\% CI Half-Width at Most 0.0038. Rank: Mean Increase of $\log(1+\text{rank})$. Modes: Mean, Global, Position, Shuffle}
\label{tab:a_grid}
\scriptsize\setlength{\tabcolsep}{3pt}
\begin{adjustbox}{max width=\textwidth}\begin{tabular}{@{}lllccc@{\hspace{14pt}}lllccc@{}}
\toprule
Block & tensor & mode & $\Delta$ acc. & log-p. drop & rank & Block & tensor & mode & $\Delta$ acc. & log-p. drop & rank\\
\midrule
L2B0 & $\Delta$ & mea & $-$0.058 & $+$0.026 & $+$0.0015 & L2B3 & $\Delta$ & mea & $-$0.016 & $+$0.005 & $+$0.0003\\
L2B0 & $\Delta$ & glo & $-$0.652$^{\dagger}$ & $+$0.034 & $+$0.0134 & L2B3 & $\Delta$ & glo & $-$0.110 & $+$0.053 & $+$0.0023\\
L2B0 & $\Delta$ & pos & $-$0.002 & $+$0.001 & $+$0.0004 & L2B3 & $\Delta$ & pos & 0.028 & $-$0.003 & $-$0.0002\\
L2B0 & $\Delta$ & shu & $-$1.170$^{\dagger}$ & $+$0.038 & $+$0.0216 & L2B3 & $\Delta$ & shu & $-$0.174 & $+$0.058 & $+$0.0026\\
L2B0 & $B$ & mea & $-$0.020 & $-$0.001 & $+$0.0003 & L2B3 & $B$ & mea & 0.028 & $+$0.003 & $-$0.0001\\
L2B0 & $B$ & glo & $-$0.242$^{\dagger}$ & $-$0.001 & $+$0.0050 & L2B3 & $B$ & glo & $-$0.200$^{\dagger}$ & $+$0.027 & $+$0.0050\\
L2B0 & $C$ & mea & $-$0.002 & $-$0.002 & $+$0.0003 & L2B3 & $C$ & mea & 0.072 & $+$0.002 & $-$0.0002\\
L2B0 & $C$ & glo & $-$0.326$^{\dagger}$ & $+$0.012 & $+$0.0074 & L2B3 & $C$ & glo & $-$0.252$^{\dagger}$ & $+$0.035 & $+$0.0063\\
L2B0 & all & mea & $-$0.150 & $+$0.025 & $+$0.0030 & L2B3 & all & mea & 0.016 & $+$0.003 & $+$0.0007\\
L2B1 & $\Delta$ & mea & 0.010 & $+$0.003 & $+$0.0002 & L3B0 & $\Delta$ & mea & $-$0.038 & $-$0.045 & $+$0.0006\\
L2B1 & $\Delta$ & glo & $-$1.188$^{\dagger}$ & $+$0.029 & $+$0.0240 & L3B0 & $\Delta$ & glo & $-$0.066 & $+$0.018 & $+$0.0007\\
L2B1 & $\Delta$ & pos & 0.020 & $+$0.001 & 0.0000 & L3B0 & $\Delta$ & pos & $-$0.018 & $-$0.028 & $+$0.0004\\
L2B1 & $\Delta$ & shu & $-$1.668$^{\dagger}$ & $+$0.042 & $+$0.0347 & L3B0 & $\Delta$ & shu & $-$0.248$^{\dagger}$ & $+$0.136 & $+$0.0033\\
L2B1 & $B$ & mea & 0.006 & 0.000 & $+$0.0006 & L3B0 & $B$ & mea & $-$0.130 & $-$0.023 & $+$0.0015\\
L2B1 & $B$ & glo & $-$0.862$^{\dagger}$ & $-$0.003 & $+$0.0171 & L3B0 & $B$ & glo & $-$0.364$^{\dagger}$ & $-$0.030 & $+$0.0050\\
L2B1 & $C$ & mea & $-$0.006 & 0.000 & $+$0.0005 & L3B0 & $C$ & mea & $-$0.106 & $-$0.028 & $+$0.0012\\
L2B1 & $C$ & glo & $-$1.116$^{\dagger}$ & $+$0.015 & $+$0.0216 & L3B0 & $C$ & glo & $-$0.240$^{\dagger}$ & $-$0.026 & $+$0.0031\\
L2B1 & all & mea & $-$0.070 & $+$0.005 & $+$0.0019 & L3B0 & all & mea & $-$0.218$^{\dagger}$ & $-$0.065 & $+$0.0025\\
L2B2 & $\Delta$ & mea & $-$0.012 & $-$0.001 & $+$0.0009 & L3B1 & $\Delta$ & mea & $-$0.102 & $-$0.033 & $+$0.0010\\
L2B2 & $\Delta$ & glo & $-$0.280$^{\dagger}$ & $-$0.005 & $+$0.0068 & L3B1 & $\Delta$ & glo & $-$0.142 & $-$0.059 & $+$0.0017\\
L2B2 & $\Delta$ & pos & 0.000 & $-$0.002 & $+$0.0004 & L3B1 & $\Delta$ & pos & $-$0.108 & $-$0.025 & $+$0.0011\\
L2B2 & $\Delta$ & shu & $-$0.220$^{\dagger}$ & $+$0.001 & $+$0.0059 & L3B1 & $\Delta$ & shu & $-$0.180 & $-$0.020 & $+$0.0026\\
L2B2 & $B$ & mea & 0.012 & 0.000 & $+$0.0002 & L3B1 & $B$ & mea & $-$0.108 & $-$0.057 & $+$0.0012\\
L2B2 & $B$ & glo & $-$0.142 & $-$0.006 & $+$0.0037 & L3B1 & $B$ & glo & $-$0.192 & $-$0.101 & $+$0.0030\\
L2B2 & $C$ & mea & $-$0.040 & $-$0.001 & $+$0.0009 & L3B1 & $C$ & mea & $-$0.028 & $-$0.076 & $+$0.0005\\
L2B2 & $C$ & glo & $-$0.108 & $-$0.007 & $+$0.0025 & L3B1 & $C$ & glo & $-$0.256$^{\dagger}$ & $-$0.104 & $+$0.0038\\
L2B2 & all & mea & $-$0.082 & $+$0.001 & $+$0.0025 & L3B1 & all & mea & $-$0.232$^{\dagger}$ & $-$0.107 & $+$0.0026\\
\bottomrule
\end{tabular}\end{adjustbox}
\end{table*}

\section{Interaction Range}
\label{app:range}
For each block of MambaVision-T we measure, on 1,000 images, how far each token draws information from. For Mamba mixers we use the exact implicit attention matrix of the selective scan (mean absolute value over channels); for attention blocks the attention maps averaged over heads. Distance is the mass-weighted grid distance between query and key tokens; for Mamba mixers we also divide by the distance of a uniform map under the same causal mask. The Mamba maps exclude the short depthwise convolution before the scan, which would make them look more local still. ERF90: number of key tokens holding 90\% of the mass.

\begin{table}[!t]
\centering
\caption{Interaction Range of MambaVision-T Blocks (Token-Grid Units, 1,000 Images; $\pm$ 95\% CI Half-Width)}
\label{tab:a_rf}
\footnotesize\setlength{\tabcolsep}{2.2pt}
\begin{adjustbox}{max width=\columnwidth}\begin{tabular}{@{}llccccc@{}}
\toprule
Block & type & distance & $\div$ uniform & vertical & horizontal & ERF90\\
\midrule
L2B0 & M & 2.70 $\pm$ 0.02 & 0.42 & 0.79 & 2.31 & 33.1\\
L2B1 & M & 3.11 $\pm$ 0.01 & 0.49 & 0.77 & 2.76 & 37.1\\
L2B2 & M & 1.92 $\pm$ 0.01 & 0.30 & 0.49 & 1.68 & 23.2\\
L2B3 & M & 3.41 $\pm$ 0.01 & 0.54 & 1.43 & 2.67 & 55.4\\
L2B4 & A & 6.24 $\pm$ 0.02 & 0.86 & 3.82 & 4.06 & 131.4\\
L2B5 & A & 6.05 $\pm$ 0.02 & 0.83 & 3.68 & 3.92 & 122.5\\
L2B6 & A & 6.60 $\pm$ 0.02 & 0.91 & 4.02 & 4.31 & 113.2\\
L2B7 & A & 7.34 $\pm$ 0.02 & 1.01 & 4.56 & 4.75 & 109.9\\
L3B0 & M & 2.48 $\pm$ 0.01 & 0.81 & 1.07 & 1.94 & 17.5\\
L3B1 & M & 2.73 $\pm$ 0.01 & 0.90 & 1.35 & 2.01 & 17.7\\
L3B2 & A & 4.12 $\pm$ 0.01 & 1.14 & 2.64 & 2.58 & 24.1\\
L3B3 & A & 4.31 $\pm$ 0.01 & 1.19 & 2.72 & 2.75 & 18.2\\
\bottomrule
\end{tabular}\end{adjustbox}
\end{table}

\section{Per-Block Content Effects Under Corruption}
\label{app:corrupt}
\Cref{tab:a_ic} gives the per-block content effects behind \cref{sec:robust}: MambaVision-T, four ImageNet-C corruptions at severity 3, first 10,000 images, clean-image reference means, ten null permutations per block.

\begin{table}[!t]
\centering
\caption{MambaVision-T Under Corruption: Content Effect $C$ (pp, Mean Null Minus Real Accuracy Change; Positive: Real Ablation Worse) per Block, 10,000 Images. $^{*}$ Holm $p<0.05$ Over 12 Blocks per Condition. Last Row: Pooled $C$ [95\% CI]}
\label{tab:a_ic}
\footnotesize\setlength{\tabcolsep}{2.2pt}
\begin{adjustbox}{max width=\columnwidth}\begin{tabular}{@{}llcccc@{}}
\toprule
Block & type & contrast & JPEG & noise & blur\\
\midrule
L2B0 & M & $-$0.14 & $-$0.75$^{*}$ & $+$4.07$^{*}$ & $-$1.11$^{*}$\\
L2B1 & M & $-$0.09 & $+$0.05 & $+$0.32 & $+$0.49$^{*}$\\
L2B2 & M & $-$0.14 & $+$0.12 & $+$0.81$^{*}$ & $+$0.12\\
L2B3 & M & $+$0.06 & $-$0.04 & $+$0.87$^{*}$ & $-$0.24\\
L2B4 & A & $-$0.21 & $-$1.53$^{*}$ & $-$4.02$^{*}$ & $-$0.99$^{*}$\\
L2B5 & A & $+$0.18 & $-$0.93$^{*}$ & $-$0.09 & $-$0.97$^{*}$\\
L2B6 & A & $+$0.02 & $-$0.02 & $+$0.02 & $-$0.44\\
L2B7 & A & $-$0.07 & $+$0.18 & $+$0.49 & $-$0.10\\
L3B0 & M & $+$0.05 & $-$0.30 & $+$0.25 & $-$0.13\\
L3B1 & M & $+$0.01 & 0.00 & $-$0.19 & $+$0.06\\
L3B2 & A & $-$0.04 & $+$0.54$^{*}$ & $+$0.25 & $+$0.54$^{*}$\\
L3B3 & A & $+$0.08 & $+$0.31 & $+$0.10 & $+$0.24\\
\midrule pooled & & $-$0.02 [$-$0.11, $+$0.07] & $-$0.20 [$-$0.31, $-$0.09] & $+$0.24 [$+$0.12, $+$0.37] & $-$0.21 [$-$0.35, $-$0.08]\\
\bottomrule
\end{tabular}\end{adjustbox}
\end{table}

\section*{Code and Data Availability}
The code for all experiments in this paper is publicly available at \url{https://github.com/tahkhan00-bit/mambavision-causal-attribution}.

\section*{Acknowledgment}
The authors used Claude (Anthropic), an AI assistant, to help write and debug the experiment and analysis code and to draft and edit the text of this manuscript. All experiments were run by the authors, all numbers were checked against the saved per-image results, and the authors take full responsibility for the content of this paper.

\bibliographystyle{IEEEtran}
\bibliography{main}

\end{document}